\documentclass[letterpaper, 10 pt, journal, twoside]{IEEEtran}
\IEEEoverridecommandlockouts 

\usepackage{hyperref, url, amsmath, amssymb, booktabs, graphicx}
\graphicspath{{images/}}

\usepackage{microtype}
\usepackage{color}

\usepackage{url}

\title{Tracking Human Pose During Robot-Assisted\\ Dressing using Single-Axis Capacitive\\ Proximity Sensing} 

\author{Zackory Erickson, Maggie Collier, Ariel Kapusta, and Charles C. Kemp%
\thanks{Manuscript received: September, 9, 2017; Revised January, 3, 2018;
Accepted February, 12, 2018.}
\thanks{This paper was recommended for publication by Editor Yasuyoshi Yokokohji upon
evaluation of the Associate Editor and Reviewers' comments. We thank C. Karen Liu and Greg Turk for their assistance with this work. This work was supported by NSF award IIS-1514258 and the National Institute on Disability, Independent Living, and Rehabilitation Research (NIDILRR), grant 90RE5016-01-00 via RERC TechSAge. Dr. Kemp is a cofounder, a board member, an equity holder, and the CTO of Hello Robot, Inc., which is developing products related to this research. This research could affect his personal financial status. The terms of this arrangement have been reviewed and approved by Georgia Tech in accordance with its conflict of interest policies.}
\thanks{Zackory Erickson, Maggie Collier, Ariel Kapusta, and Charles C. Kemp are with the Healthcare Robotics Lab, Georgia Institute of Technology, Atlanta, GA., USA. (email: {\small zackory@gatech.edu}; {\small collierm@uab.edu}; {\small akapusta@gatech.edu}; {\small charlie.kemp@bme.gatech.edu})}%
\thanks{Digital Object Identifier (DOI): see top of this page.}
}

\begin{document}

\maketitle

\begin{abstract}
Dressing is a fundamental task of everyday living and robots offer an opportunity to assist people with motor impairments. While several robotic systems have explored robot-assisted dressing, few have considered how a robot can manage errors in human pose estimation, or adapt to human motion in real time during dressing assistance. In addition, estimating pose changes due to human motion can be challenging with vision-based techniques since dressing is often intended to visually occlude the body with clothing. We present a method to track a person's pose in real time using capacitive proximity sensing. This sensing approach gives direct estimates of distance with low latency, has a high signal-to-noise ratio, and has low computational requirements. Using our method, a robot can adjust for errors in the estimated pose of a person and physically follow the contours and movements of the person while providing dressing assistance. As part of an evaluation of our method, the robot successfully pulled the sleeve of a hospital gown and a cardigan onto the right arms of 10 human participants, despite arm motions and large errors in the initially estimated pose of the person's arm. We also show that a capacitive sensor is unaffected by visual occlusion of the body and can sense a person's body through cotton clothing.
\end{abstract}


\section{INTRODUCTION}
\label{sec:intro}

\IEEEPARstart{R}{obotic} assistance provides an opportunity to increase independence and privacy for people with motor impairments, such as some older adults, who need assistance with dressing.
However, providing assistance with dressing remains a challenging task for robots.

Most robot-assisted dressing approaches consider the task of pulling a garment onto a human or mannequin body at a known pose~\cite{shinohara2011learning,gao2016iterative,chance2016assistive}, yet few have explored how a robot can correct for errors in the estimated pose of a person, or track human motion during dressing. Estimated pose error can result from either poor initial estimation or movement of the body after the pose was estimated. In addition, dressing often aims to visually occlude a person's body with a garment and the robot's arms can further occlude the body. Because of this, human pose estimation during dressing and tracking pose changes due to human motion is difficult for standard vision-based approaches~\cite{gao2015user}.




\begin{figure}
\centering
\includegraphics[width=0.48\textwidth, trim={15cm 8cm 6cm 1cm}, clip]{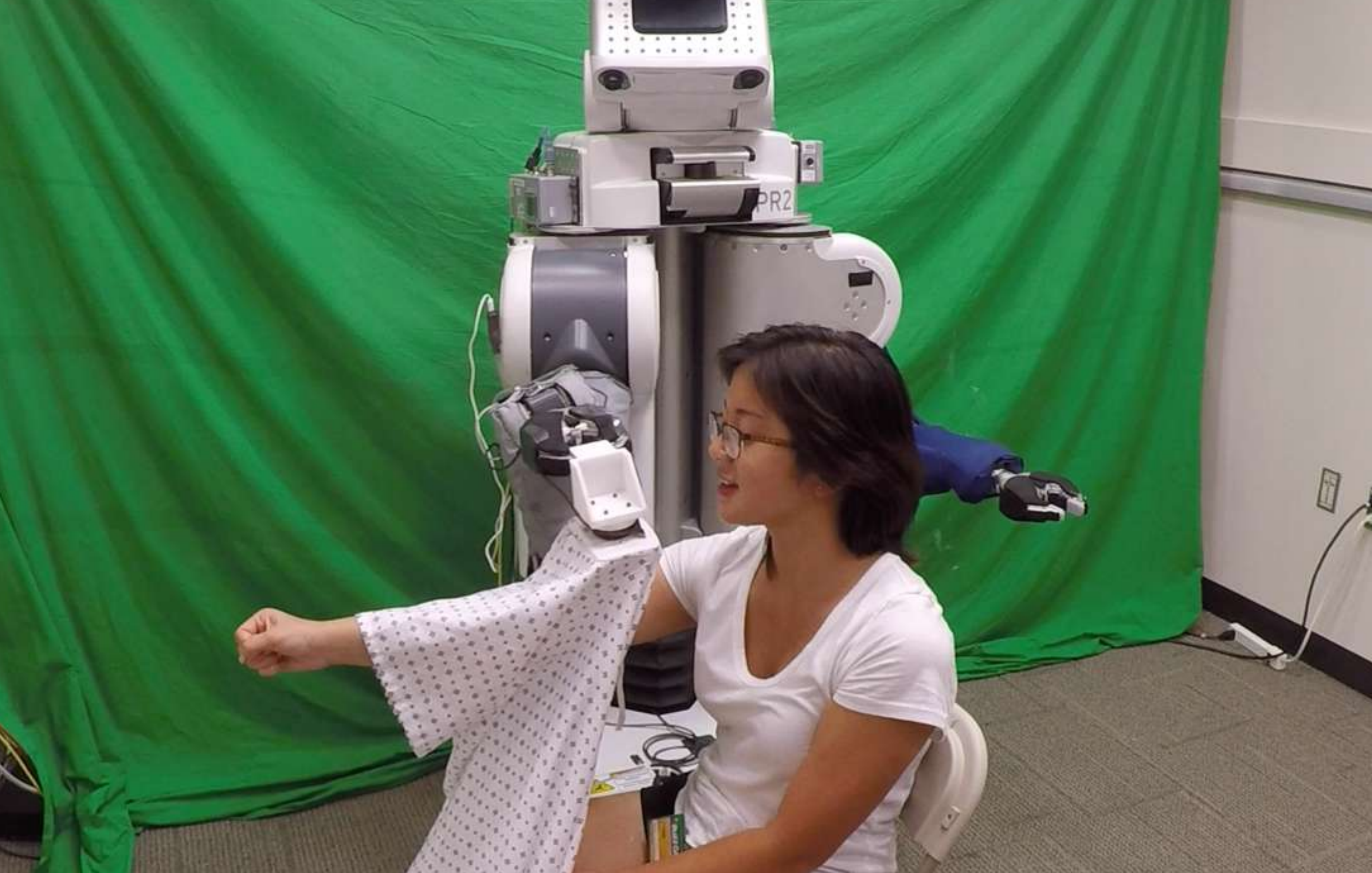}
\caption{\label{fig:intro}A PR2 pulling a hospital gown onto a human participant's arm. The robot uses our capacitive sensing method to adjust for pose estimation errors and human movement during dressing.}
\end{figure}

In this work, we present a method that uses capacitive sensing to estimate distance to a person's arm in real time, enabling a robot to follow the contours of the person's arm and track the person's motion during robot-assisted dressing. Our capacitive sensing approach can directly estimate the distance to a person's arm using a closed form equation which we derive from the capacitance equation for a standard parallel plate capacitor. Unlike many computer vision approaches for pose estimation, our approach has minimal computational requirements due to the low-dimensional capacitance signal. Furthermore, this approach is rapid, with low latency and high sampling rate, and has a high signal-to-noise ratio. These qualities allow a robot to be highly responsive to both error and human motion during robot assistance when using just a simple PD controller.

We evaluated our method with 10 human participants in a study during which a robot pulled the sleeve of both a hospital gown and a knitted sweater (e.g., a cardigan) onto the participant's arm, as shown in Fig.~\ref{fig:intro}. We show that a robot can use our capacitive sensing method to adapt to errors in human pose estimation and successfully dress a person's arm despite large error. We also demonstrate how our method allows a robot to track human motion even when a person's arm is already covered by fabric clothing. In addition, capacitive sensing can also be used to detect when contact occurs between a robot and the person receiving assistance, which we discuss in Section~\ref{sec:movement}. From our evaluation, we found that capacitance measurements had low variability across all participants, which suggests that our capacitive sensing approach may generalize well when providing assistance to a wider population of people.


In this paper, we make the following contributions:
\begin{itemize}
\item We propose and evaluate a capacitive sensing approach for regulating the distance between a robot's end effector and a person's body.
\item We show that our capacitive sensing approach can enable a robot to follow the contours of a person's arm even during unscripted arm movement.
\item We provide evidence that the capacitance to distance measurements are consistent across people and do not require individual calibration.
\item We demonstrate that this approach is able to follow the contours of a person's arm even when the person is already wearing cotton clothing that may cause visual occlusions, as can occur when assisting with an outer layer of clothing.
\end{itemize}


\section{RELATED WORK}
\label{sec:related_work}


\subsection{Robot-Assisted Dressing}

Several researchers have considered robot-assisted dressing, yet almost all previous work assumes that a person will hold a stationary pose while being dressed. In addition, few have considered how a robot can adjust for pose estimation errors that occur after dressing begins. We observe this to be an important issue facing prior robot assistance research for other assistive tasks, where human users would often change their pose between when the pose was estimated and the task was completed~\cite{park2017multimodal}.
One area of work has considered the use of visual information to determine a person's initial pose before dressing. Klee et al. used a vision system to detect a person's pose before a Baxter robot assisted in putting on a hat~\cite{klee2015personalized}. Their vision system ensured that a person held the correct stationary pose and was used to halt robot movement if human motion was detected.

Another research area has focused on visual features for tracking the state of a garment during dressing. Koganti et al. used RGB-D and motion capture data to estimate the topological relationship between a garment and a person's body~\cite{koganti2017bayesian}.
Tamei et al. used a motion capture system to estimate the topological relationship between a garment and a stationary mannequin~\cite{tamei2011reinforcement}. They then had a dual-arm robot learn to pull a T-shirt over the mannequin's head using reinforcement learning. Similarly, Yamazaki et al. used a depth camera to determine a trajectory for a humanoid robot to assist in pulling up a pair of trousers and they used optical flow to estimate the cloth state~\cite{yamazaki2014bottom}. However, these works all assumed the person remains stationary while being dressed.

Pignat et al. used a Baxter robot to pull one sleeve of a jacket onto a person's arm~\cite{pignat2017learning}. The robot tracked the person's hand movement in real time using an AR tag and the robot began dressing once the person placed his or her hand in the sleeve entry of the jacket. They did not explicitly model nor account for errors or human motion after the robot began dressing a person. Similarly, Chance et al. used a Baxter robot to determine dressing errors and clothing types when dressing a human participant~\cite{chance2017quantitative}. They varied the initial pose of the human participants. Their robot followed a predetermined trajectory based on the person's initial pose and they did not consider limb movement during dressing. Erickson et al. used a physics-based simulation to estimate the forces applied to a person's body during robot-assisted dressing using haptic and kinematic measurements at the robot's end effector~\cite{erickson2017does}. Their system accounts for translation and rotation of the human limb before dressing, but does not model human motion during dressing assistance. In previous work, Kapusta et al. observed that some starting heights of linear trajectories along the arm may fail and rapidly cause high forces on the person's arm~\cite{kapusta2016data}. In this work, we show that our method enables a robot to avoid some failure cases by adapting in real time to the estimated pose of the arm without applying high forces.

Unlike most past work, which has assumed a person remains stationary when being dressed, our research focuses on sensing and adapting to human motion during dressing assistance. Related to human motion and pose estimation, Gao et al. used a Baxter robot to assist a person in putting on a sleeveless jacket, wherein the person would push their arms through the sleeves~\cite{gao2015user}. Despite this, the robot followed a fixed trajectory given by the person's initial pose which was estimated using a top-view depth camera. As the researchers noted, \textit{the depth sensor could not be used to determine a person's pose during dressing because the robot's arms would block visual sight of the body}. Similarly, Gao et al. has considered how a robot can adjust for human movement during dressing using force feedback control~\cite{gao2016iterative}. They used RGB-D data to estimate a person's pose and compute an initial robot trajectory for dressing a sleeveless jacket. They then proposed a stochastic path optimization approach for personalized robot-assisted dressing that locally adjusts the robot's motion based on force feedback from the robot's end effector. However, force feedback allows a robot to react to human motion only \textit{after} the robot begins applying forces to the person's body. We propose a capacitive sensing approach that enables a robot to directly estimate human motion in real time and track a person's movement before contact is made or forces are applied.

\subsection{Capacitive Sensing}

\begin{figure*}
\centering
\vspace{8pt}
\includegraphics[width=0.48\textwidth, trim={7cm 3.5cm 7cm 3.1cm}, clip]{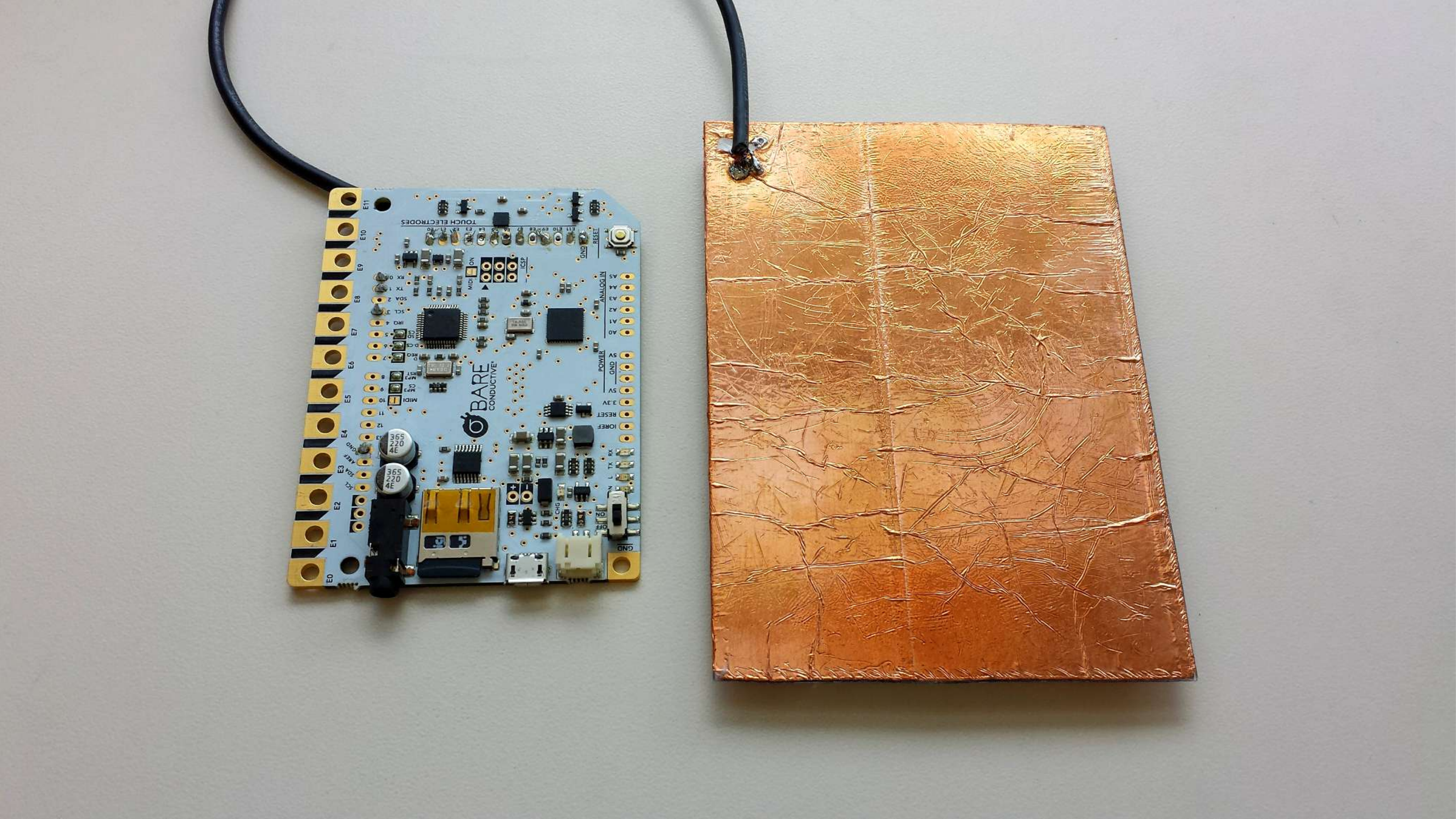}
\includegraphics[width=0.48\textwidth, trim={0.5cm 1cm 2.11cm 0cm}, clip]{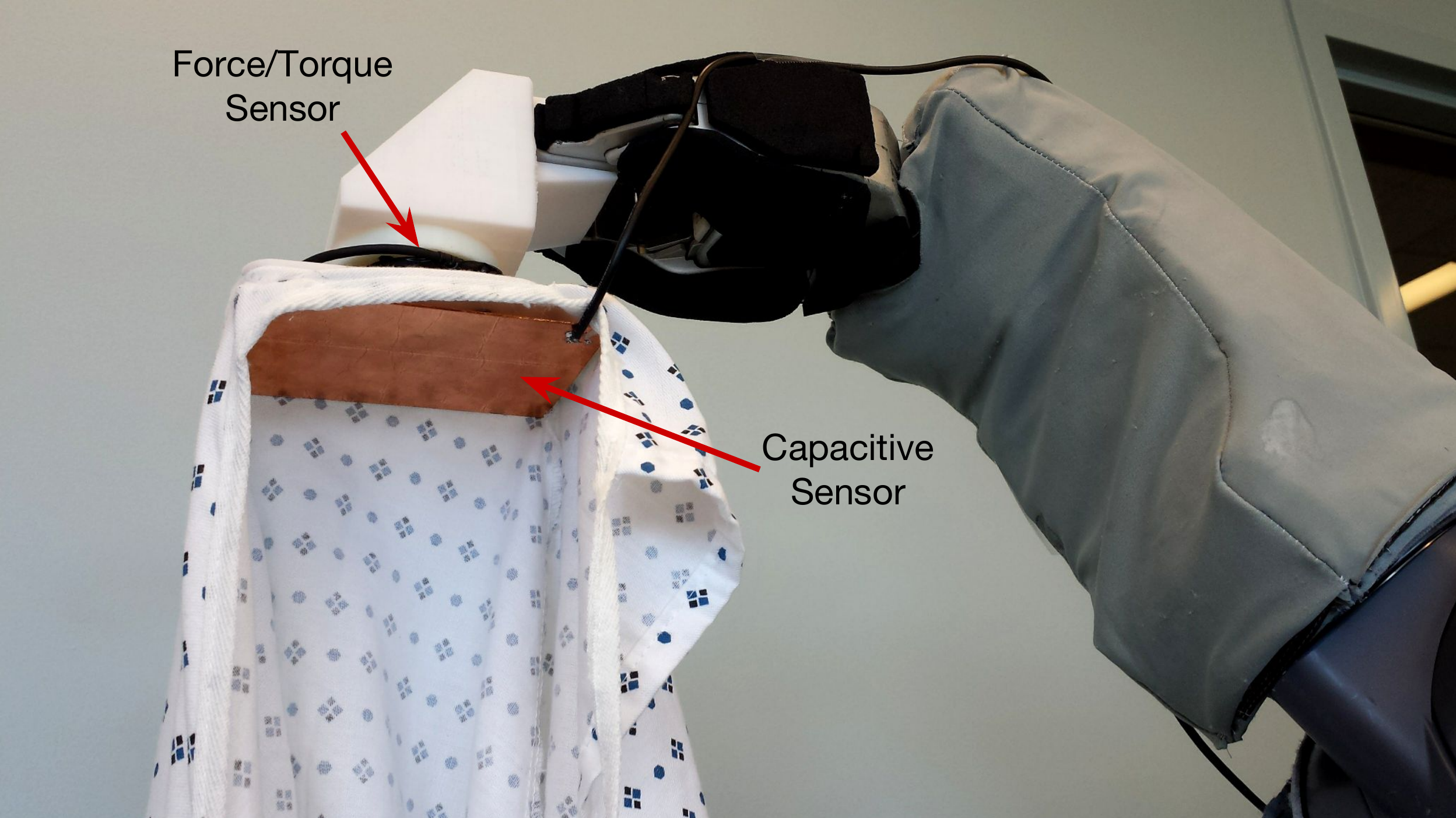}
\caption{\label{fig:sensors}Left: Copper plate capacitive sensor and the Bare Conductive microcontroller. Right: The capacitive sensor is mounted to the bottom of a tool that encases an ATI force/torque sensor and holds the hospital gown.}
\end{figure*}

Capacitive sensors have been widely used in robotics for contact-based tactile sensing applications. For example, work by Muhammad et al. used capacitive sensing for texture recognition by scanning over a surface and comparing frequency spectrums of sensor measurements~\cite{muhammad2011capacitive}. Another area of work has focused on integrating capacitive sensors within artificial skins for obtaining tactile measurements upon contact~\cite{damilano2017robust, cotton2009multifunctional, ulmen2010robust}. These tactile measurements can include information such as applied pressure, force, or strain. Schmitz et al. developed a capacitive skin for contact pressure sensing and they implemented their sensors onto the bodies of three humanoid robots~\cite{schmitz2011methods}. Phan et al. used a capacitive sensing robotic skin for impact monitoring and force reduction~\cite{phan2011capacitive}. In addition, Ji et al. designed a capacitive robotic skin for capturing high sensitivity force measurements~\cite{ji2016design}. In comparison to contact-based sensing, we focus on proximity sensing to measure the distance between a robotic end effector and a person's body.


Capacitive proximity sensors have been broadly explored within human-robot interaction. For example, capacitive proximity systems have been implemented as safety mechanisms that enable a robot to avoid collisions with humans and objects~\cite{xia2016multi, schlegl2013virtual, hoffmann2016environment}. Kirchner et al. used capacitive sensors mounted on a robotic arm for both proximity sensing and material recognition by also comparing measurements taken at different sensor drive frequencies~\cite{kirchner2008capacitive}. Lee et al. designed a capacitive sensor that provides both tactile and proximity sensing for artificial robotic skin~\cite{lee2009dual}. The researchers used a 16 $\times$ 16 grid of capacitive sensors to detect a human hand from 17 cm away. When compared to materials such as metal, plastic, and wood, the researchers found that a human hand resulted in the largest capacitance change as distance between the object and sensor decreased. 

In work that closely relates to ours, Navarro et al. proposed the use of capacitive proximity sensing for safer human-robot interaction~\cite{navarro2013methods}. The researchers designed a large 3 by 16 grid of capacitive sensors for determining the location and proximity of a human hand with respect to the sensor. Unlike our work, they used a stationary table-mounted sensor with multiple grid cells to sense hand movement, whereas we use a single sensor cell mounted on a robot's end effector to sense and react to human arm motion in real time. Navarro et al. then demonstrated that 2 $\times$ 2 arrays of capacitive sensors mounted on a robot's end effector allowed the robot to follow the curvature of objects, such as aluminum and wooden rods~\cite{navarro20146d}. Unlike our work, the researchers used two parallel capacitive plates, each with four sensors, and assumed a stationary object remained between the two plates. Our work relies on only a single capacitive plate to sense and adjust for human motion during robotic assistance, and we derive an equation for estimating the distance to a person's arm which is analogous to the parallel plate capacitor model.



\section{METHOD}
\label{sec:method}

\subsection{Capacitive Sensor Design}

To create the sensor's electrode, we covered one side of an 11.5~cm $\times$ 8.5~cm $\times$ 1~mm acrylic sheet in copper foil tape. We connected the electrode to a Bare Conductive Touch Board microcontroller\footnote{Bare Conductive: \url{https://www.bareconductive.com/}}, a commercially available board that is designed for filtering readings from a capacitive sensor for touch and proximity-based applications. This board uses a MPR121 capacitive proximity sensor controller that includes high and low frequency noise filtering. We are able to sample capacitance measurements from this sensor at over 200 Hz. We used nonconductive adhesive to mount the electrode to the bottom of a tool held by the PR2's end effector, and we secured the microcontroller to the forearm of the PR2. This tool contains an ATI force/torque sensor and allows the robot to more easily hold a garment. The PR2 is a general-purpose mobile manipulator from Willow Garage with two 7-DoF back-drivable arms and an omni-directional mobile base. Fig.~\ref{fig:sensors} shows the sensor by itself and mounted on the PR2. In future iterations, these capacitive electrodes could be implemented directly into the front or backside of a robot's fingertip. The lack of sensitivity to cloth between the sensor and a human body, which we explore in Section~\ref{sec:movement}, also presents other possibilities for electrode placement.


\begin{figure}
\centering
\includegraphics[width=0.48\textwidth, trim={1cm 17.9cm 2cm 3cm}, clip]{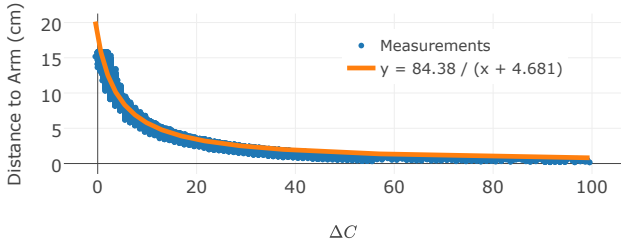}
\caption{\label{fig:calibration}Capacitance to distance function fit to sample data along all six human arm locations. The PR2's end effector followed a downward vertical trajectory until contact with six locations evenly spaced along a human arm. The X-axis represents the capacitance change from a baseline defined when no object is underneath the sensor. As discussed in Section~\ref{sec:captodist}, we found that this function consistently fit capacitance measurements from all 10 participants in our experimental evaluation.}
\end{figure}

\subsection{Estimating Distance to a Human Arm}
\label{sec:captoprox}

Given measurements from the capacitive sensor during dressing, our method estimates a distance between the robot's end effector and a person's arm. To accomplish this, we first collected capacitance measurements when the capacitive sensor and the robot's end effector were at various distances from a person's arm. Specifically, we had the robot position its end effector at six equally spaced locations along the person's arm, from fist to shoulder. At each location, the robot's end effector would start 15 cm above the person's arm and move 1 cm/s downward while we recorded capacitance readings and positions based on the robot's forward kinematics at 100 Hz. Fig.~\ref{fig:calibration} displays the resulting graph of distance against capacitance measurements. Empirically, we found that our capacitive sensor had a maximum sensing range of approximately 10~cm from a human arm. Capacitance readings taken at a distance further than 10~cm from the arm appeared to be mostly indistinguishable from one another. 

We fit a function to this data to estimate the distance between the robot's end effector and a person's arm given a capacitance measurement taken during dressing. The function we fit is inspired by the capacitance equation for a parallel plate capacitor which is defined as,


\begin{equation}
\label{eq:capacitance}
C = \frac{k\epsilon_0A}{d}
\end{equation}
where $C$ is capacitance, $k$ is the relative permittivity of the dielectric between the plates, $\epsilon_0 = 8.85 \times 10^{-12}$ is the vacuum permittivity, $A$ is the overlapping surface area of the plates, and $d$ is the distance between the plates. We set $\alpha=k\epsilon_0 A$ and reorganize (\ref{eq:capacitance}) as,

\begin{equation}
\label{eq:capacitance2}
d = \frac{k\epsilon_0A}{C} = \frac{\alpha}{C}
\end{equation}

We defined and fit the following function, which is analogous to (\ref{eq:capacitance2}) and is defined as,

\begin{equation}
\label{eq:captodist}
d(\Delta C) = \frac{\alpha}{\Delta C+\beta}
\end{equation}
where $\Delta C$ is the change in capacitance from a baseline defined when no object is underneath the sensor. We added a constant $\beta$ to the denominator to allow for horizontal translation of the curve with respect to $\Delta C$. We fit this function to the data shown in Fig.~\ref{fig:calibration} using least squares optimization. This resulted in the constants $\alpha=84.38$ and $\beta=4.681$, which achieved a coefficient of determination (R$^2$) of 0.969. As discussed in Section~\ref{sec:captodist}, capacitance versus distance measurements were nearly identical for all 10 human participants, and our evaluation results indicate that this function was accurate for estimating distance to all participants' arms.

\subsection{Control}

We implemented a Cartesian controller on the PR2, which uses the Orocos Kinematics and Dynamics Library\footnote{KDL: \url{http://www.orocos.org/kdl}} to provide joint-level input to the PR2's low-level PID controllers. We ran the Cartesian controller at 10~Hz. For participant safety, the robot's arms were compliant and we used low gains for all arm joints. Additionally, we ran a force threshold monitor that halted all robot movement if forces measured at the robot's end effector exceeded 10~N. We designated the X~axis as running along the arm and the Z~axis as pointing upward, opposite gravity. At each time step the controller commands the robot's end effector to move $0.5$~cm in the X~direction, towards the person's shoulder. It additionally commands the robot's end effector to raise or lower in the Z direction with respect to the error in distance from the arm. We can represent our PD controller operating in the Z direction as,

\begin{equation}
 \begin{aligned}
 u_{z}(t)  =  K_p e(t) + K_d \frac{de(t)}{dt}\\
  \end{aligned} \label{eq:controller}
\end{equation}
where
 \begin{equation}
 \begin{aligned}
  e(t)  &=   d_{desired} - d_{measured}\\
    &=   d_{desired} -  d(\Delta C)
   \end{aligned} \label{eq:error}
\end{equation}

$K_p$ and $K_d$ are the proportional and derivative gains, respectively. $e(t)$ is the tracking error: the error between the desired and measured distance from the arm at time $t$. We manually tuned the controller to be responsive to human motion, selecting $K_p = 0.3$, $K_d = 0.2$, and $d_{desired}=5\text{ cm}$.

\subsection{Design Limitations}

Our approach currently faces a few limitations. To improve sensitivity, our capacitive electrode does not have a ground plate or an active shield, thereby leaving the sensor unshielded from external sources of electromagnetic interference (EMI). Certain clothing types might also impact capacitance measurements, such as those with metal threads or metal decorations. In addition, with only a single large capacitive sensor, we can only estimate the person's arm location along a single direction. Our method has been successful at tracking human motion when error has remained primarily in a single direction, but alterations would be needed to account for pose estimation error and arm movement along multiple directions. For example, sensing 3D human motion, including vertical, lateral, and rotation movements, will likely require three capacitive sensors. Further signal processing, such as low-pass filtering, and learning techniques may also be necessary to resolve interdependencies between multiple nearby sensors.

\section{EVALUATION}

We conducted a study with 10 participants with approval from the Georgia Institute of Technology Institutional Review Board (IRB), and obtained informed consent from all participants.

We recruited able-bodied participants to meet the following inclusion/exclusion criteria: $\geq18$ years of age; have not been diagnosed with ALS or other forms of motor impairments; and fluent in written and spoken English. Of the 10 participants, 3 were female and 7 were male. Their ages ranged from 21 to 27 years and their arm lengths ranged from 65~cm to 76~cm.

Every participant started each trial seated comfortably while holding his or her arm in a specified posture, namely: arm straight, pointed forward, parallel to the floor and perpendicular to the front of the robot base; fingers curled into a fist; knuckles vertically aligned. Fig.~\ref{fig:intro} shows this configuration. We recorded the height of a participant's fist and provided this height visually as a reference point before each trial so that the participant could position his or her arm at a consistent height. A supplementary video of our experiments is available online\footnote{Video: \url{http://healthcare-robotics.com/cap-prox}}.

\begin{figure}
\centering
\includegraphics[width=0.48\textwidth, trim={1cm 17.9cm 2cm 1.5cm}, clip]{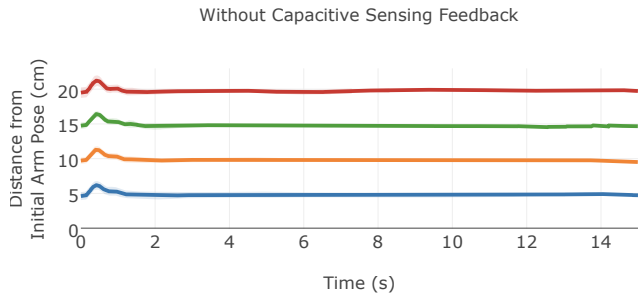}
\caption{\label{fig:position} Distance of the robot's end effector from the person's arm during dressing, estimated from PR2 forward kinematics and starting height of the person's fist. Colored lines represent trials with a different initial distance between a participant's hand and the robot's end effector before dressing begins. Results are averages from 50 dressing trials: 5 trials at each height for each of the 10 participants. When capacitive sensing feedback is not used, the robot follows a predefined straight path. There is 1.2~cm ($\pm$0.5~cm) of error that occurred at the start of each trial, which may be attributable to the low gains of the PR2 and the outstretched arm at the start of a trial.}
\end{figure}

\begin{figure}
\centering
\includegraphics[width=0.48\textwidth, trim={0cm 0cm 0cm 0cm}, clip]{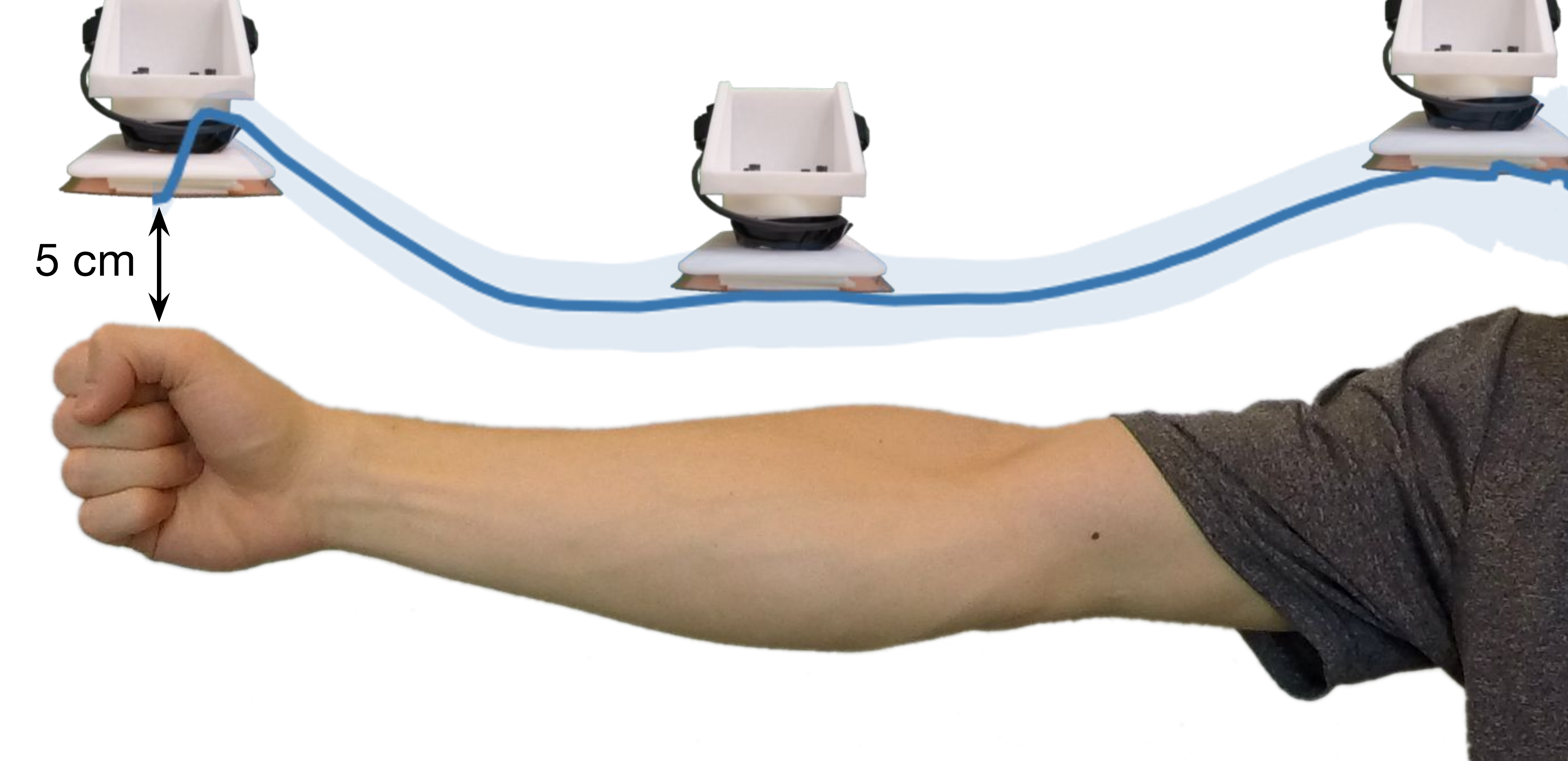}
\caption{\label{fig:contours} With our method, the robot's end effector is able to follow the contours of a person's arm during dressing. We overlay the average movement trajectory taken by the robot when using capacitive sensing feedback to dress all 10 participants. Background shading represents one standard deviation.
}
\vspace{-8pt}
\end{figure}



\subsection{Error in Human Pose Estimation}

As detailed in Section \ref{sec:related_work}, past work in robot-assisted dressing has relied on visual pose estimation to estimate how a person's arm is oriented prior to pulling on a jacket. However, a person's true arm pose may differ from what was initially estimated, due to visual pose estimation error, modeling error, human movement after the arm pose was estimated, or other factors. We show that our approach using feedback control with capacitive sensing can enable a robot to adapt to even large errors of 15~cm in the estimated pose of a person's arm.

We compared our approach against the robot performing open-loop movements for which the robot did not adjust the vertical height of its end effector using capacitive sensing. When dressing a human participant, the robot's end effector started at a predetermined height above the person's hand and then followed a trajectory approximately parallel to the axis of the person's arm in order to pull on a hospital gown. For each participant, the PR2 performed 10 dressing trials at four predetermined heights: 5~cm, 10~cm, 15~cm, and 20~cm above the person's fist, for a total of 40 dressing trials. We used $d_{desired}=5$~cm, so the initial pose estimation error for the four heights were 0~cm, 5~cm, 10~cm, and 15~cm respectively. For half of the 40 trials, the PR2 moved the gown in a linear trajectory along the axis of the arm without feedback from the capacitive sensor. For the other half, the robot used capacitive sensing to measure and adjust for the vertical offset between its end effector and the person's arm. When using the capacitive sensor, the robot aimed to keep its end effector 5 cm above the person's arm throughout the entire dressing trial. We alternated methods for the 10 trials at each height.

During each dressing trial, the robot measured the forces and torques at its end effector, the position of its end effector, and capacitance readings. The robot halted all movement if the magnitude of the total force measured by the ATI force/torque sensor exceeded 10 N. We collected all measurements at a frequency of 100 Hz. Fig.~\ref{fig:position} shows the vertical position of the robot's end effector during the dressing trials when capacitive feedback is not used. Note that the reported positions are with respect to the initial height of the robot's end effector, which is offset from the top of a participant's fist, as seen in Fig.~\ref{fig:contours}. For cases in which the robot's end effector began 5 or 10~cm above a participant's fist, the open-loop trajectory succeeded in pulling the sleeve of the gown up a person's arm. When starting 15~cm or 20~cm above the hand, the open-loop trajectory often led to the arm getting caught in the sleeve, or the arm missing the sleeve entirely.



\begin{figure}
\centering
\includegraphics[width=0.48\textwidth, trim={1cm 16.5cm 2cm 1.5cm}, clip]{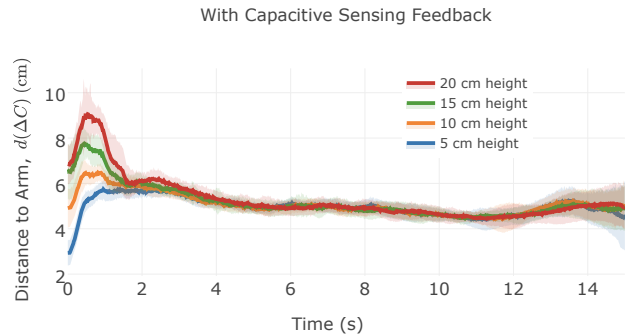}
\caption{\label{fig:proximity}Estimated distance measurements between the capacitive sensor and a human participant arm. Distance results are computed as shown in~{(\ref{eq:captodist})} and are averaged across 50 dressing trials for each starting height. Shaded regions represent one standard deviation.
}
\vspace{-8pt}
\end{figure}

In comparison, when using capacitive sensing, the end effector was able to adjust for pose estimation errors by moving downward, closer to the participant's arm. Fig.~\ref{fig:proximity} presents the estimated distance and standard deviation between the end effector and a participant's arm when using capacitive feedback control. These distance estimates are averaged over all 10 participants and were computed from capacitive sensor measurements as shown in~(\ref{eq:captodist}). The figure shows that distance estimates between the robot's end effector and a participant's arm remain between 4 to 6~cm for most of a dressing trial. Furthermore, these capacitive sensing estimates enable the robot to track the contours of a human arm, which is depicted in Fig.~\ref{fig:contours}. However, estimates are less accurate at distances exceeding the 10~cm maximum sensing range and at the start of the trials. Error at the start of trials may be due to electromagnetic interference (EMI) between the sensor cable and the robot that occurs when the robot's arm is fully extended outwards. Active shielding may help reduce these sources of interference in the future. As shown in Fig.~\ref{fig:position}, a bump in the trajectory also occurred at the start of each trial, which may be attributable to the low gains of the PR2 and its outstretched arm at the start of each trial.


\subsection{Generalization of Capacitance Measurements}
\label{sec:captodist}

We presented an optimized function in Section~\ref{sec:captoprox} to estimate the distance between the robot's end effector and a person's arm, given a capacitive sensor measurement. This function was fit to measurements taken from a single person's arm. In this section, we investigate the variance in capacitance measurements across different people and we evaluate how well the optimized function generalizes across individuals.

\begin{figure}
\centering
\includegraphics[width=0.48\textwidth, trim={1cm 17.5cm 2cm 3cm}, clip]{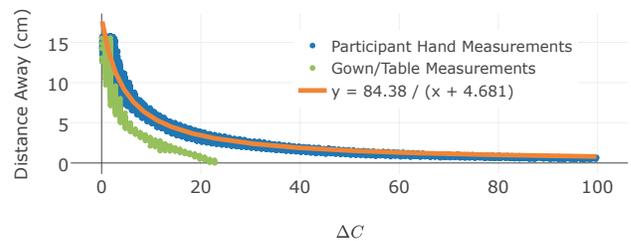}
\caption{\label{fig:calibration_participants}Capacitance measurements taken when the PR2's end effector followed a vertical trajectory above each participant's hand. This process is similar to that presented in Fig.~\ref{fig:calibration}, yet is performed with all 10 participants to evaluate variance in capacitance measurements across people. We overlay the function fit to data from Fig.~\ref{fig:calibration}, which achieves an R$^2$ value of 0.964 on measurements from all 10 participants. The green curve represents measurements captured when the capacitive sensor makes contact with a hospital gown that was resting on a plastic table.}
\vspace{-8pt}
\end{figure}

For each participant, we measured capacitance signals when the robot's end effector started 15~cm above the participant's hand and moved 1~cm/s downward until contact and then back upward at 1~cm/s. Fig.~\ref{fig:calibration_participants} displays these capacitance measurements for all 10 participants. The standard deviation in distance between the robot's end effector and a participant's arm is less than 0.2~cm when averaged across all capacitance measurements. Furthermore, we overlay the optimized function from Fig.~\ref{fig:calibration} which achieves an R$^2$ value of 0.964 and is nearly identical to the R$^2$~=~0.969 achieved when the function was fit to measurements from a single person. Overall, we found that capacitance measurements and the optimized function were consistent across participants.

Kirchner et al. and Lee et al. found that capacitance measurements taken over the human body are easily identifiable from many other materials found in human environments~\cite{kirchner2008capacitive,lee2009dual}. To further validate this, we collected measurements when the capacitive sensor made contact with a fabric gown that was resting on a plastic table. These measurements of both the table and gown are depicted in Fig.~\ref{fig:calibration_participants} by the green curve. For distances less than 10~cm, these measurements are noticeably different from measurements over a human arm, and this aligns with findings by both research groups.


\begin{figure*}
\centering
\vspace{8pt}
\includegraphics[width=0.19\textwidth, trim={15cm 8cm 6cm 1cm}, clip]{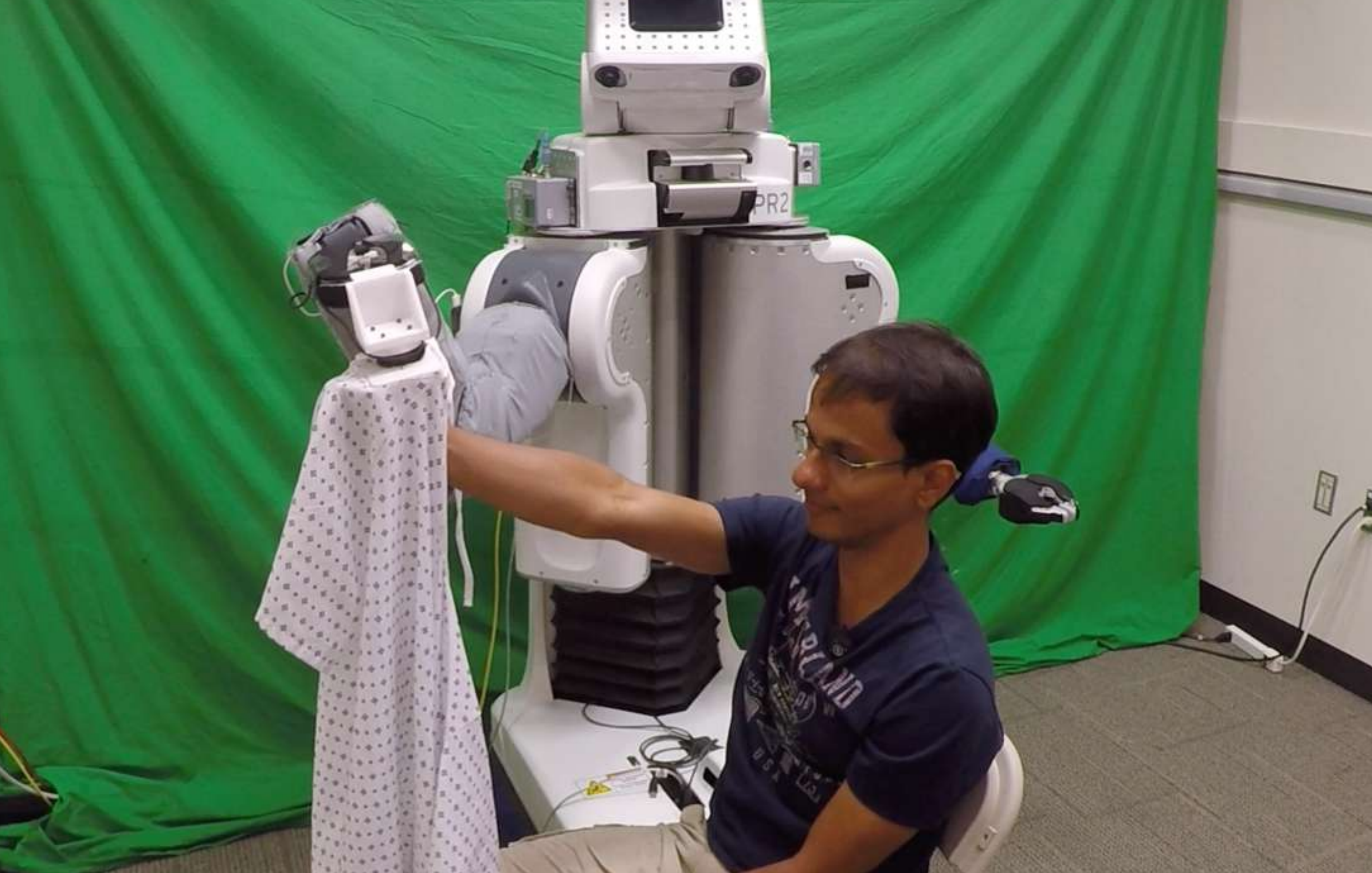}
\includegraphics[width=0.19\textwidth, trim={15cm 8cm 6cm 1cm}, clip]{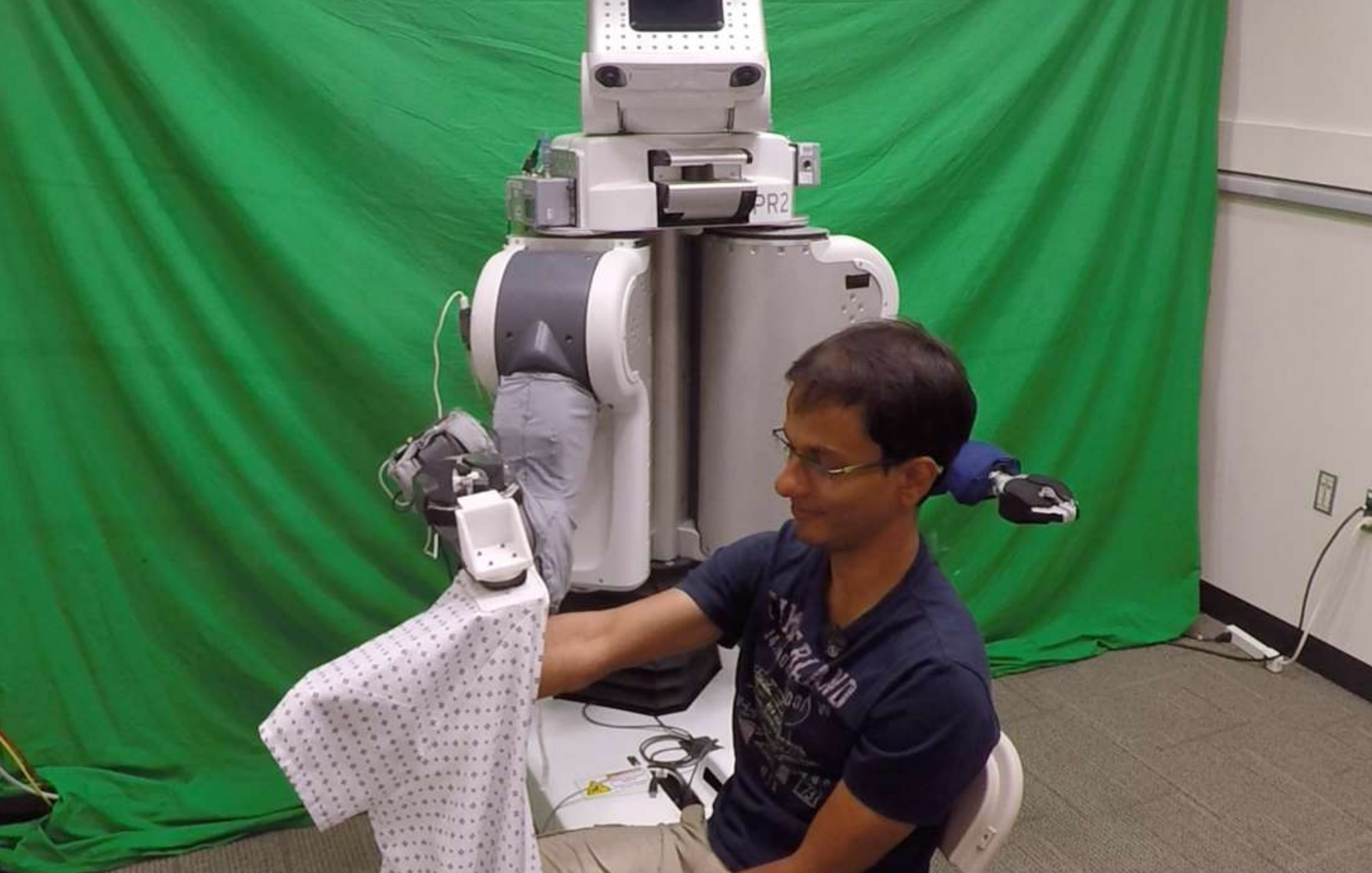}
\includegraphics[width=0.19\textwidth, trim={15cm 8cm 6cm 1cm}, clip]{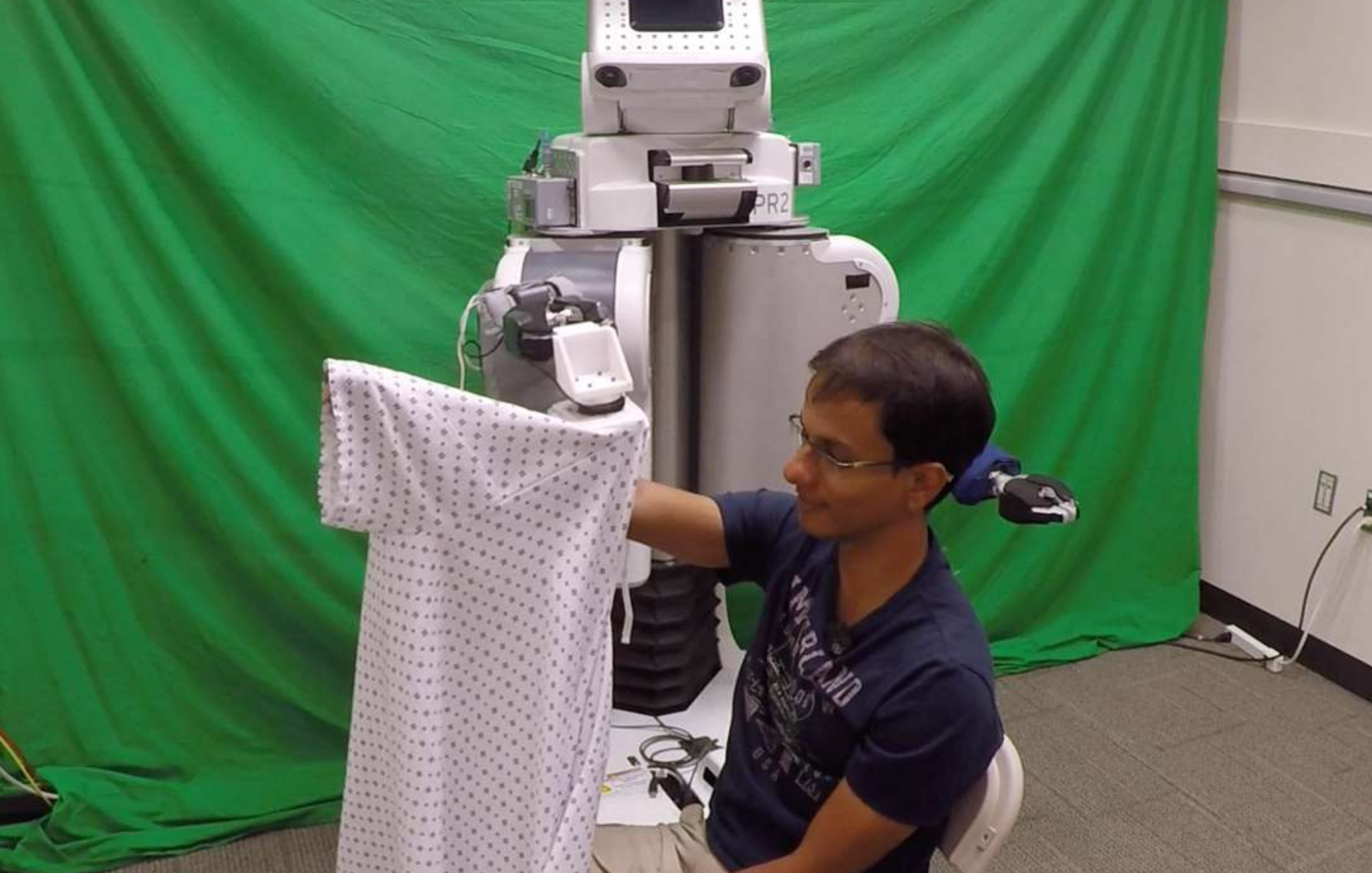}
\includegraphics[width=0.19\textwidth, trim={15cm 8cm 6cm 1cm}, clip]{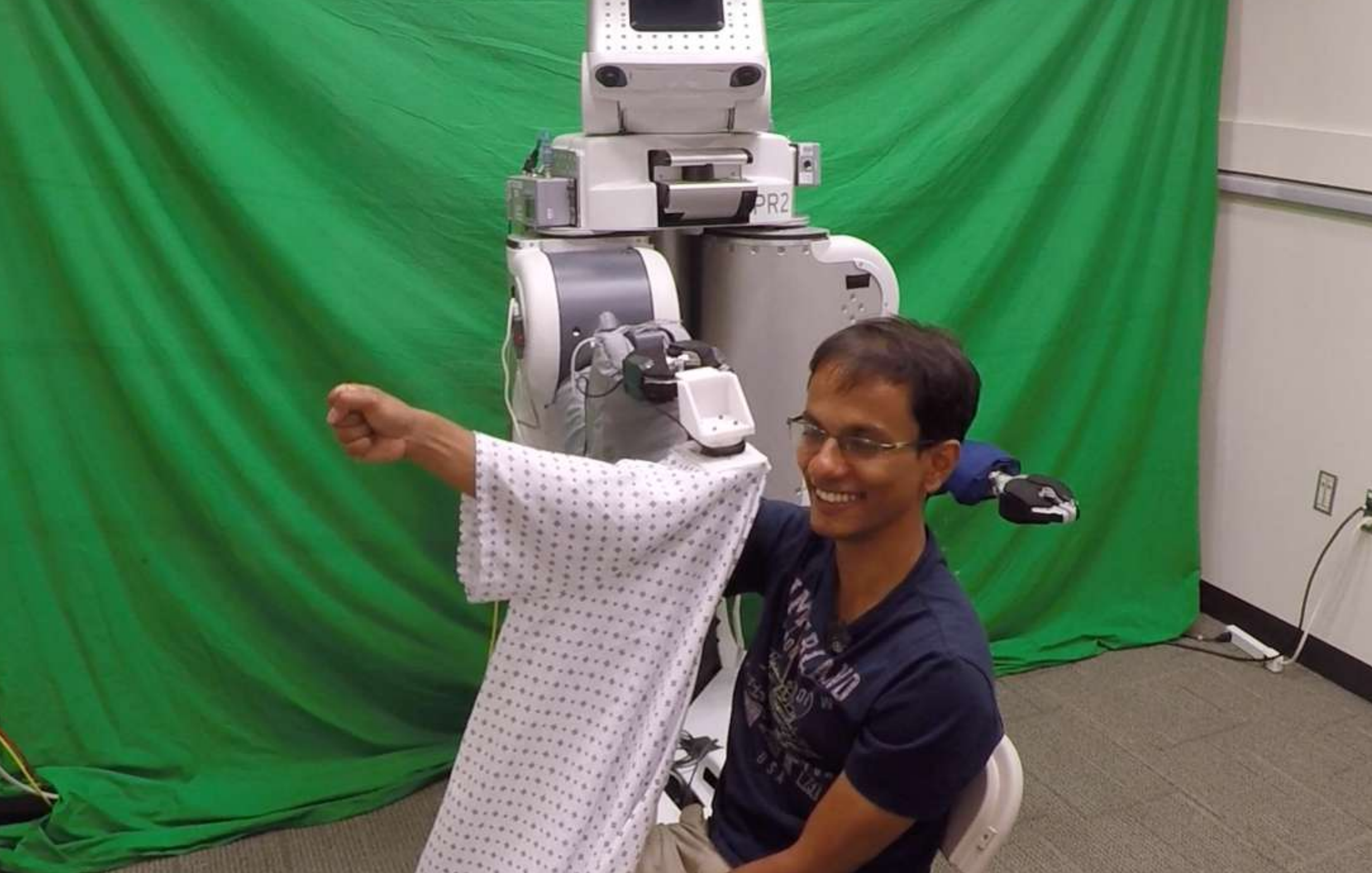}
\includegraphics[width=0.19\textwidth, trim={15cm 8cm 6cm 1cm}, clip]{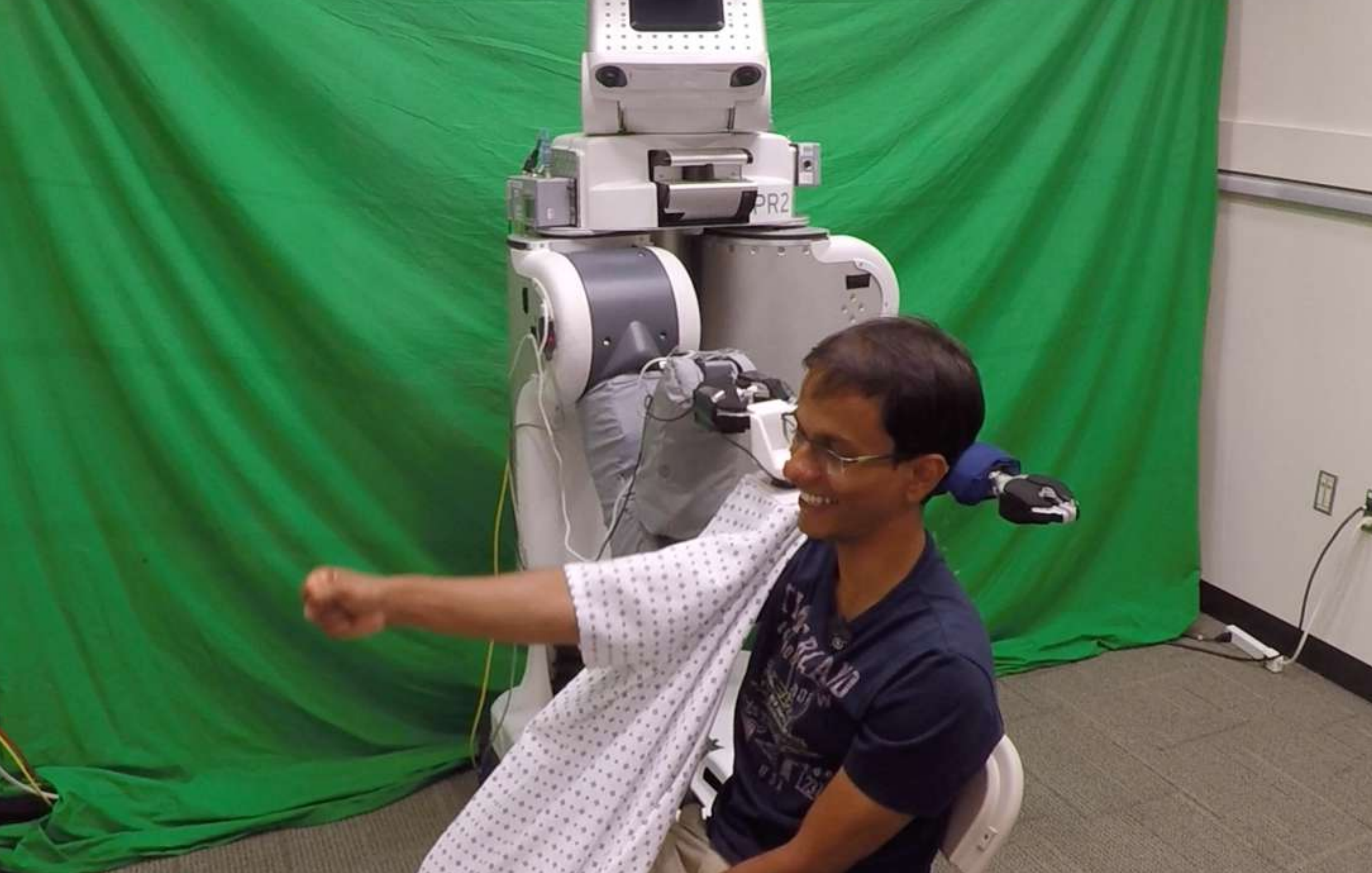}
\caption{\label{fig:demo}The PR2 uses our method to track human movement in real time during dressing assistance with a hospital gown.}
\end{figure*}

\begin{figure}
\centering
\includegraphics[width=0.48\textwidth, trim={1cm 18cm 2cm 3.5cm}, clip]{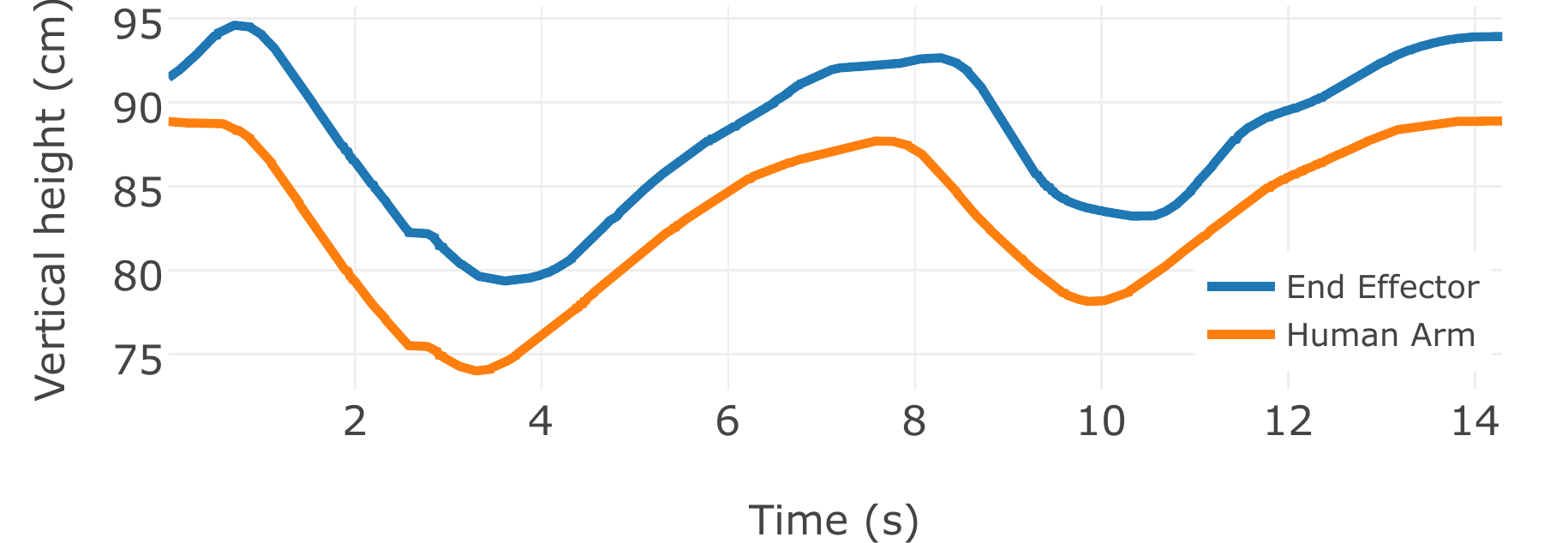}
\caption{\label{fig:stage8}Vertical position of the robot's end effector and human arm as the participant moves his/her arm during dressing. This plot shows a representative trial from a randomly selected participant during the gown dressing scenario, as seen in Fig.~\ref{fig:demo}. The vertical position of the robot's end effector is measured from forward kinematics while the vertical height of a participant's arm is estimated from capacitance measurements.}
\end{figure}

\begin{figure}
\centering
\includegraphics[width=0.48\textwidth, trim={1cm 18cm 2cm 3.5cm}, clip]{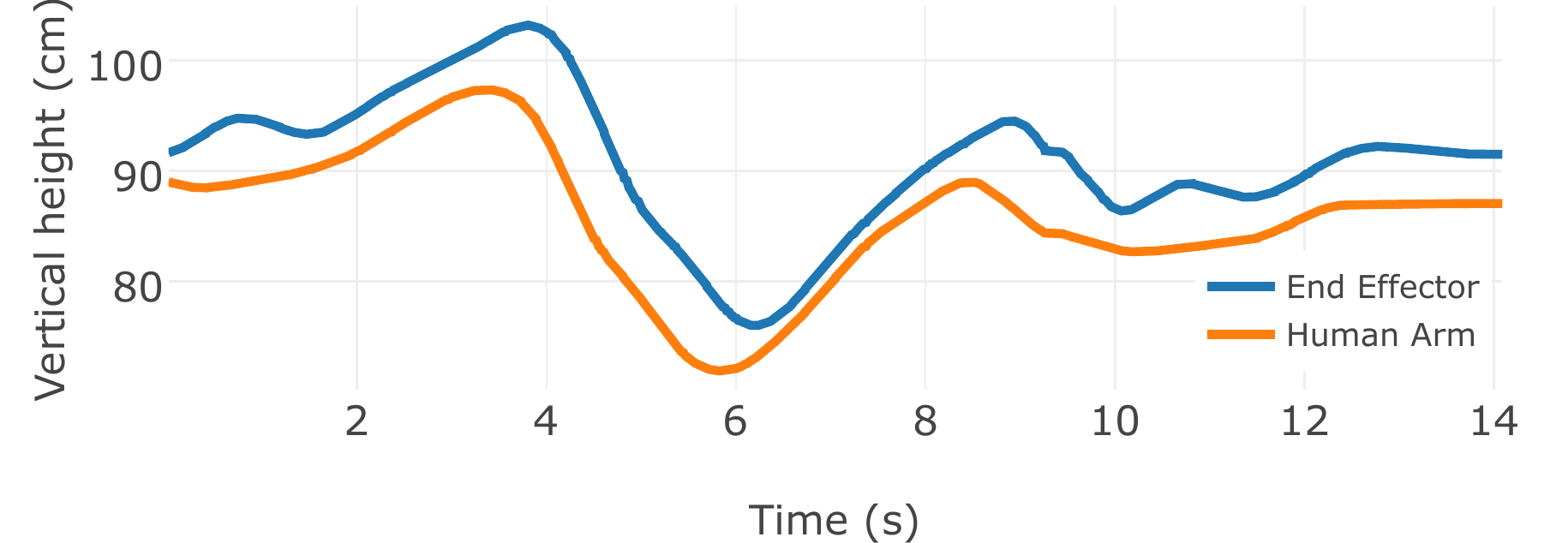}
\caption{\label{fig:stage9}Vertical position of the robot's end effector and human arm as the participant moves his/her arm while wearing a long sleeve shirt. This plot shows a representative trial from a single participant during the gown dressing scenario seen in Fig.~\ref{fig:longsleeve}. We observe that the capacitive sensor is able to track a person's arm movement, regardless of whether the person's arm is covered by clothing.}
\end{figure}

\begin{figure*}
\centering
\vspace{8pt}
\includegraphics[width=0.19\textwidth, trim={15cm 8cm 6cm 1cm}, clip]{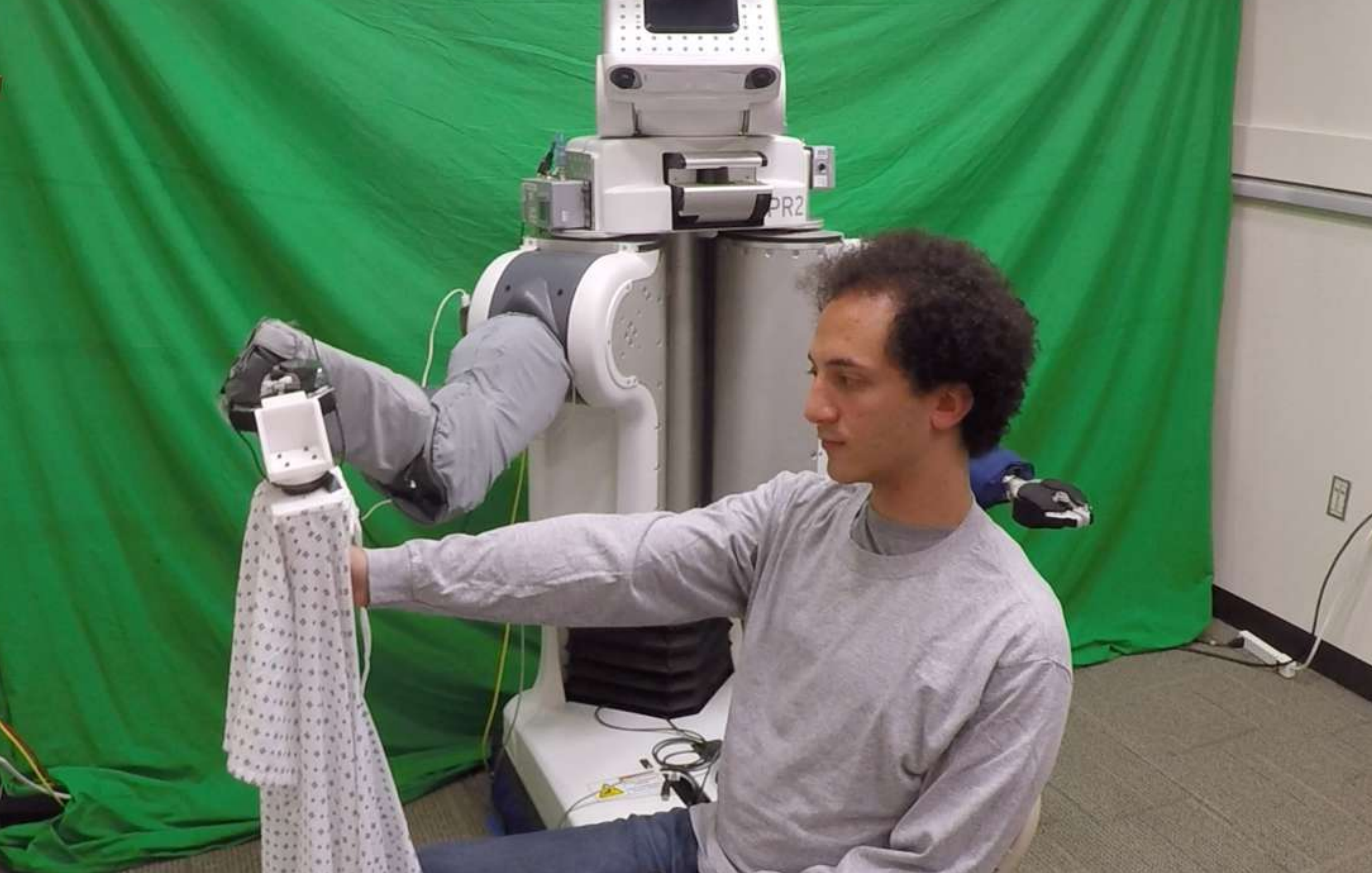}
\includegraphics[width=0.19\textwidth, trim={15cm 8cm 6cm 1cm}, clip]{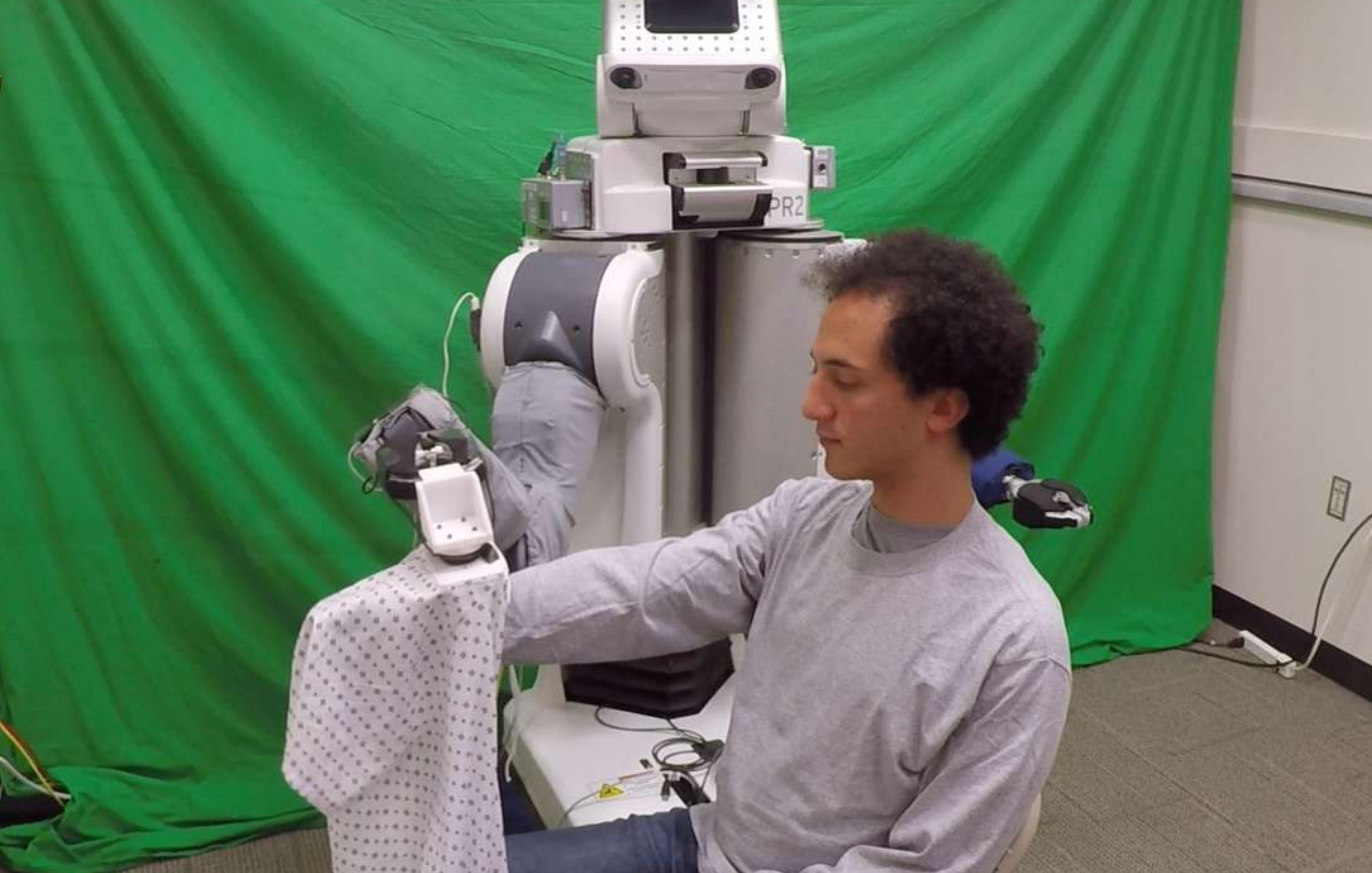}
\includegraphics[width=0.19\textwidth, trim={15cm 8cm 6cm 1cm}, clip]{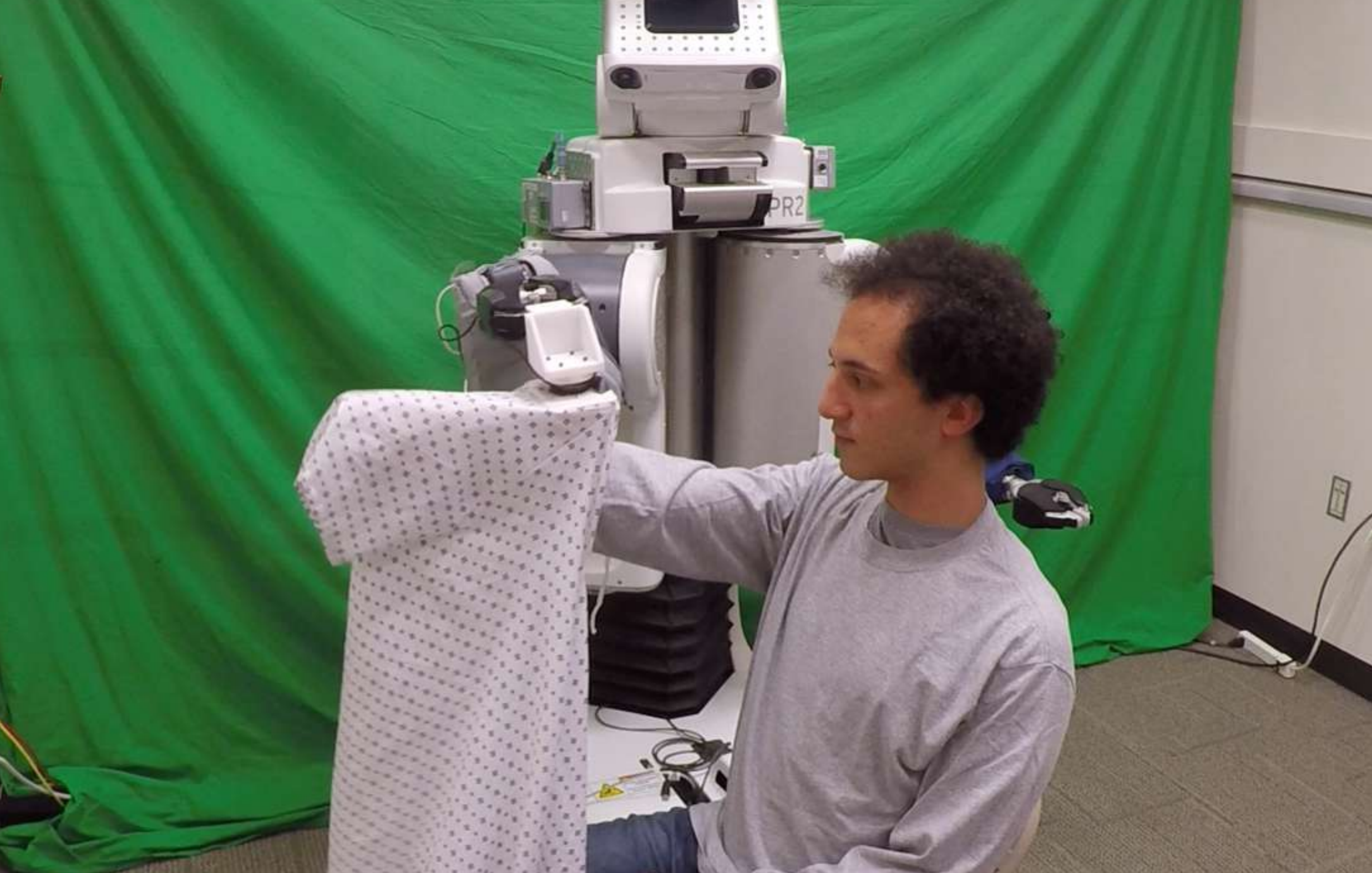}
\includegraphics[width=0.19\textwidth, trim={15cm 8cm 6cm 1cm}, clip]{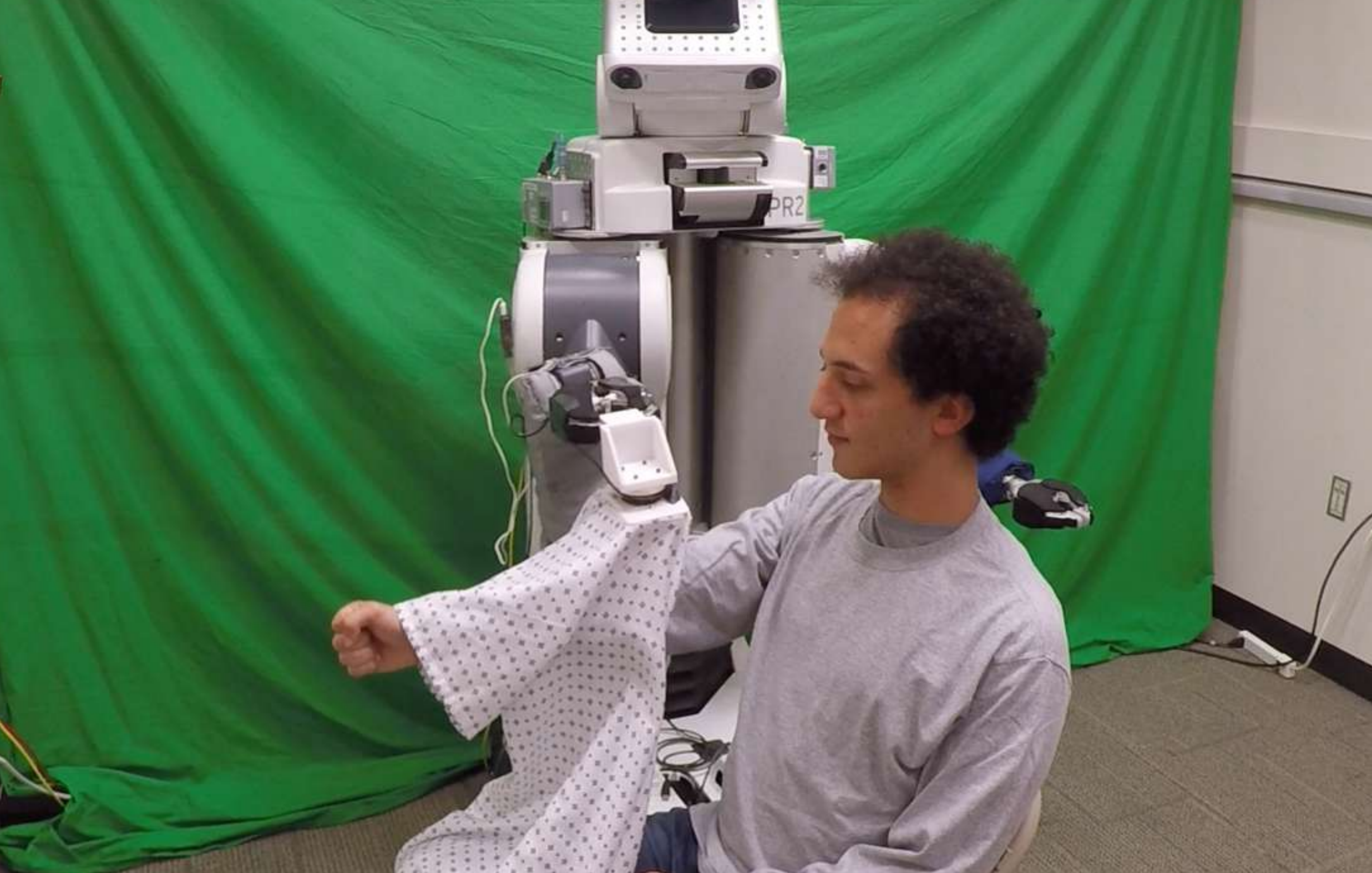}
\includegraphics[width=0.19\textwidth, trim={15cm 8cm 6cm 1cm}, clip]{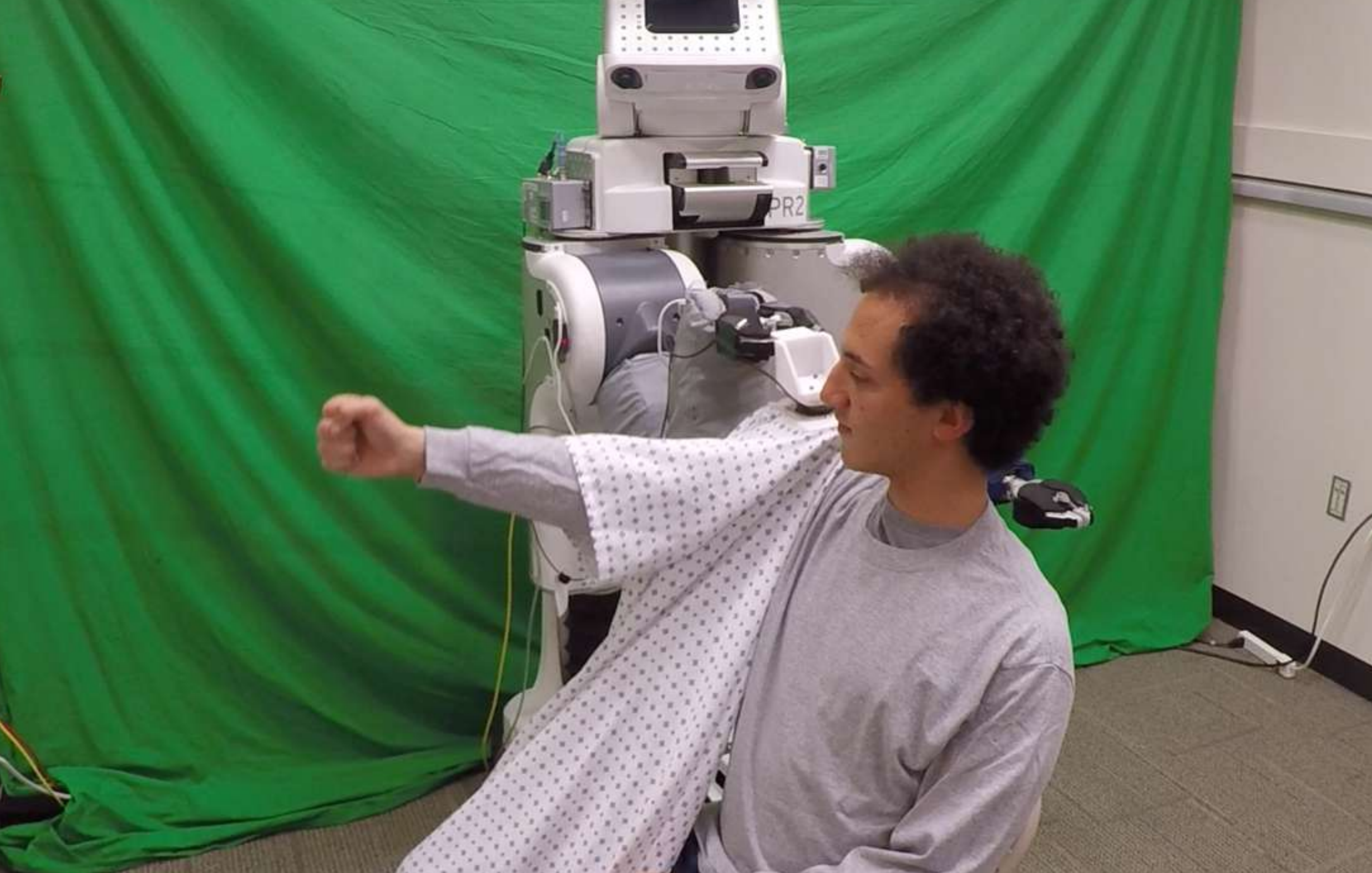}
\caption{\label{fig:longsleeve}The capacitive sensor tracking a person's arm through clothing. The PR2 is able adapt to a participant's movement even when the participant is already wearing a long sleeve cotton shirt.}
\end{figure*}

\begin{figure*}
\centering
\includegraphics[width=0.19\textwidth, trim={15cm 8cm 6cm 1cm}, clip]{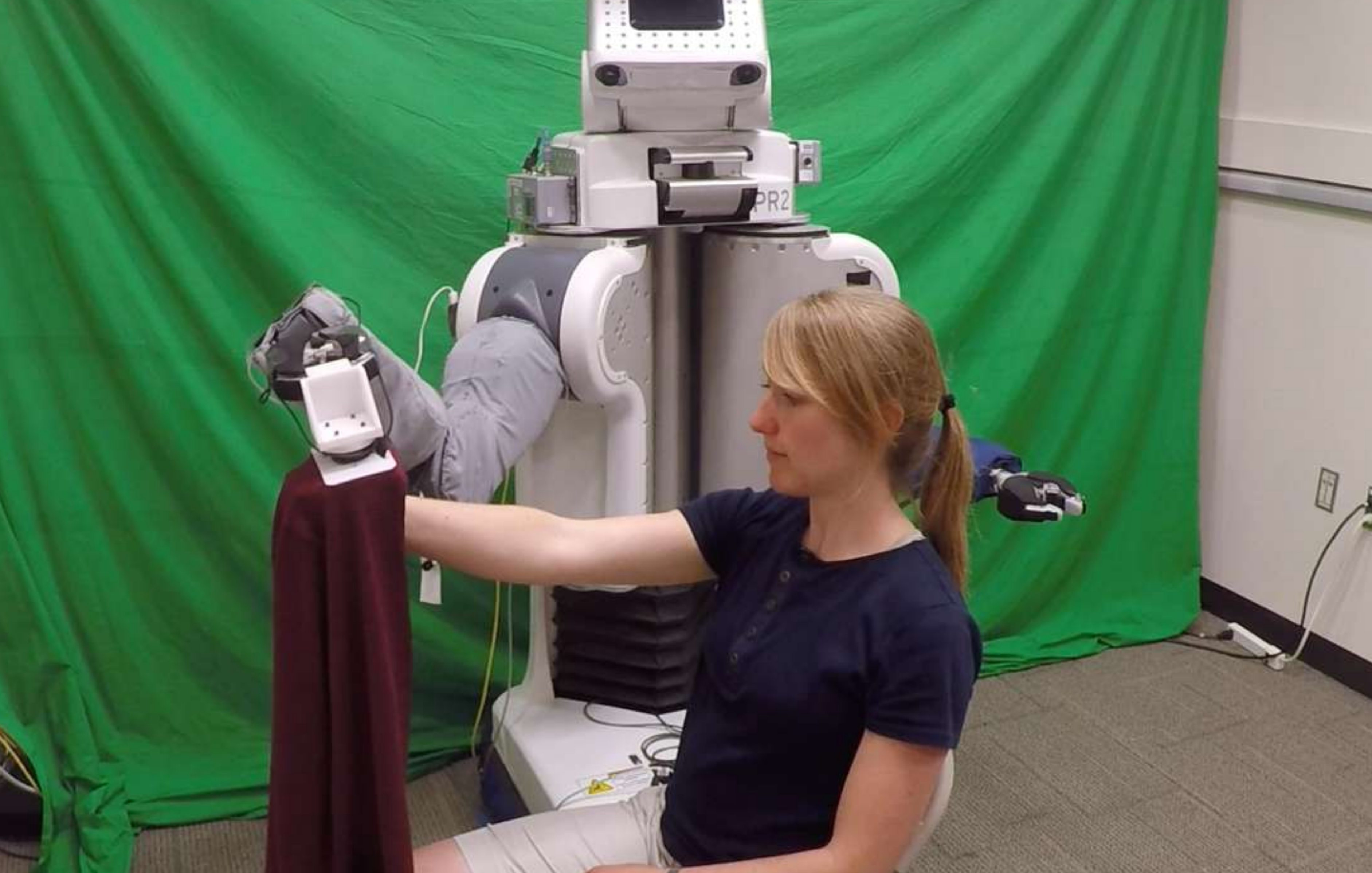}
\includegraphics[width=0.19\textwidth, trim={15cm 8cm 6cm 1cm}, clip]{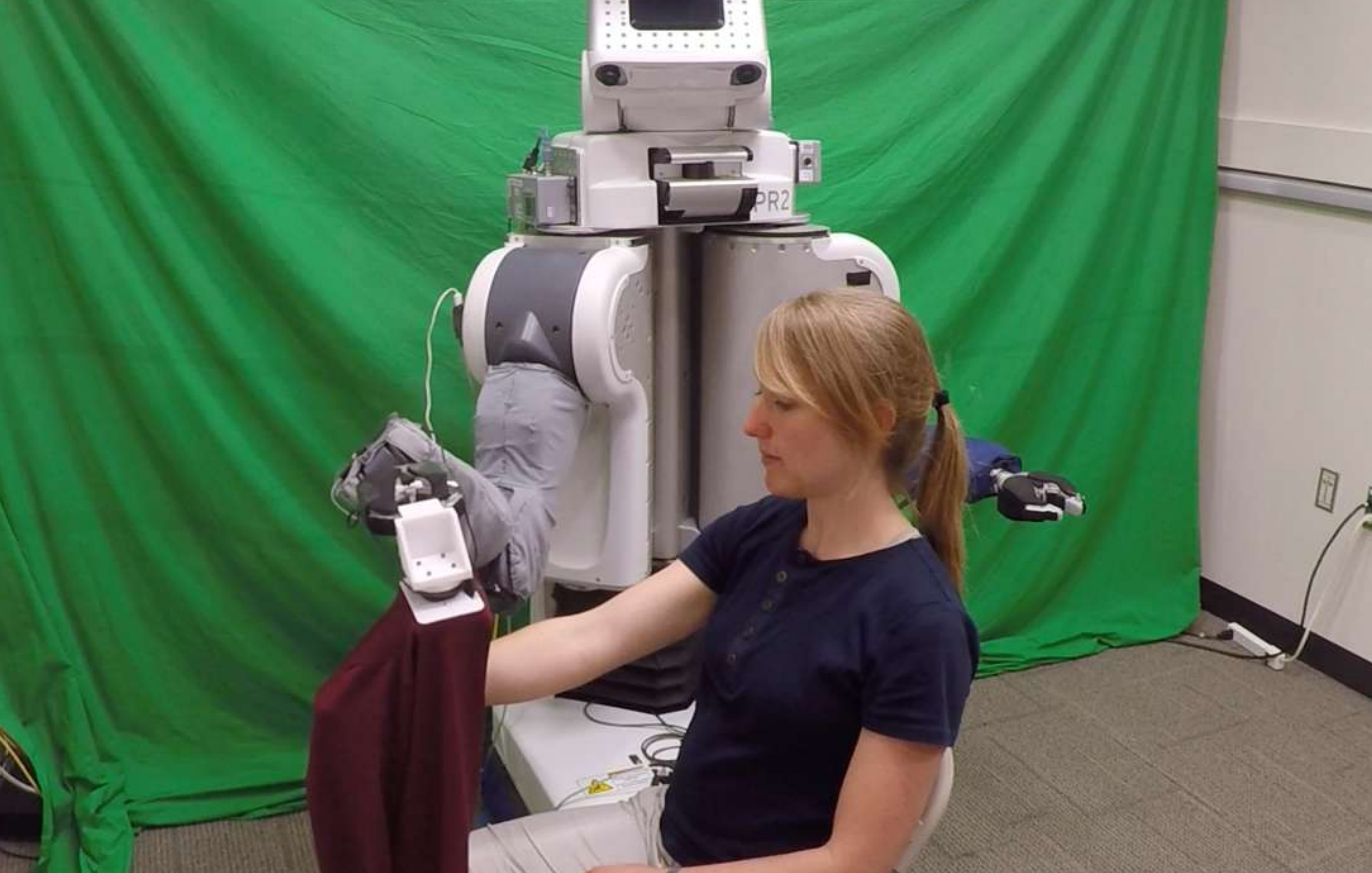}
\includegraphics[width=0.19\textwidth, trim={15cm 8cm 6cm 1cm}, clip]{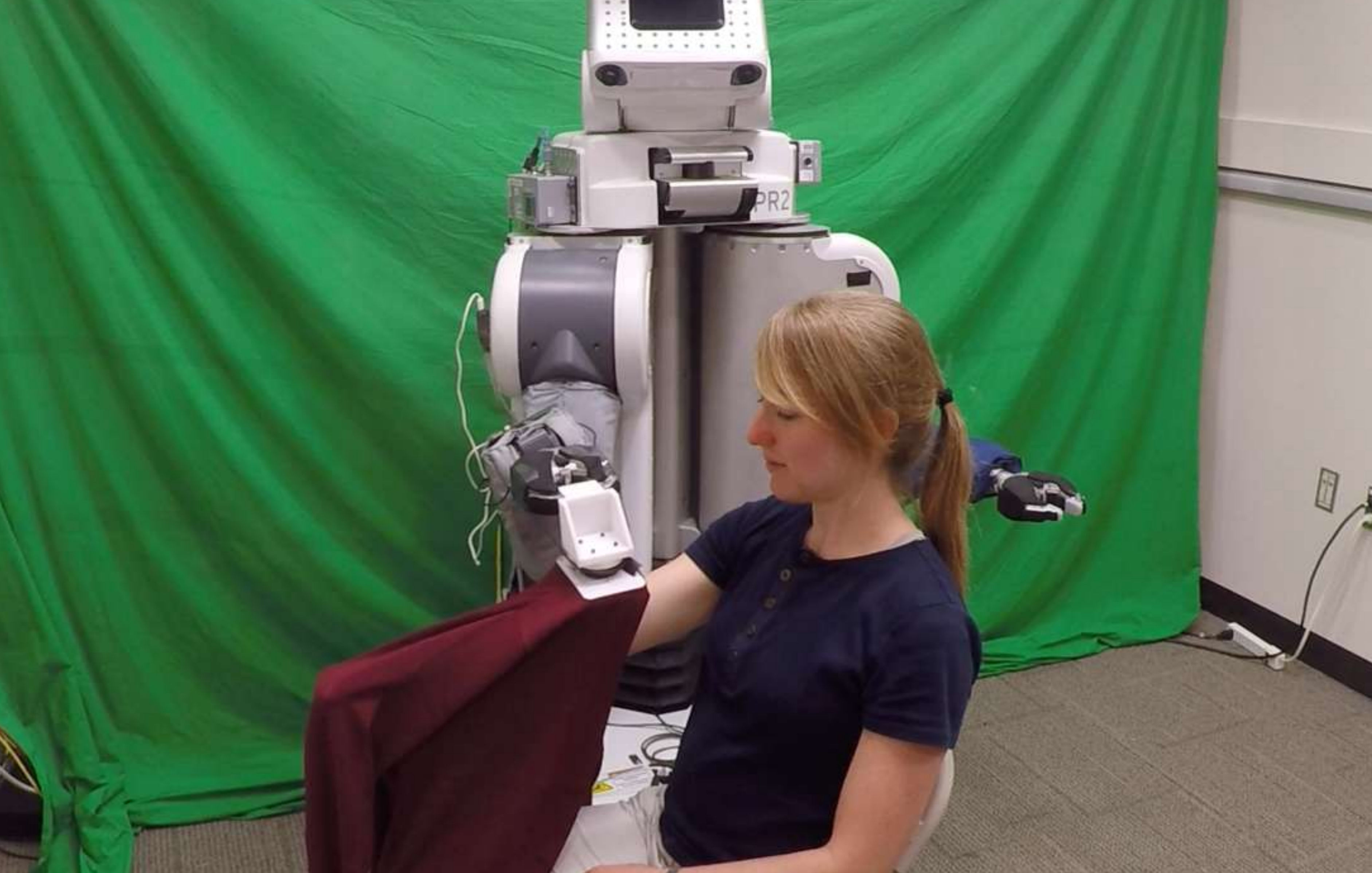}
\includegraphics[width=0.19\textwidth, trim={15cm 8cm 6cm 1cm}, clip]{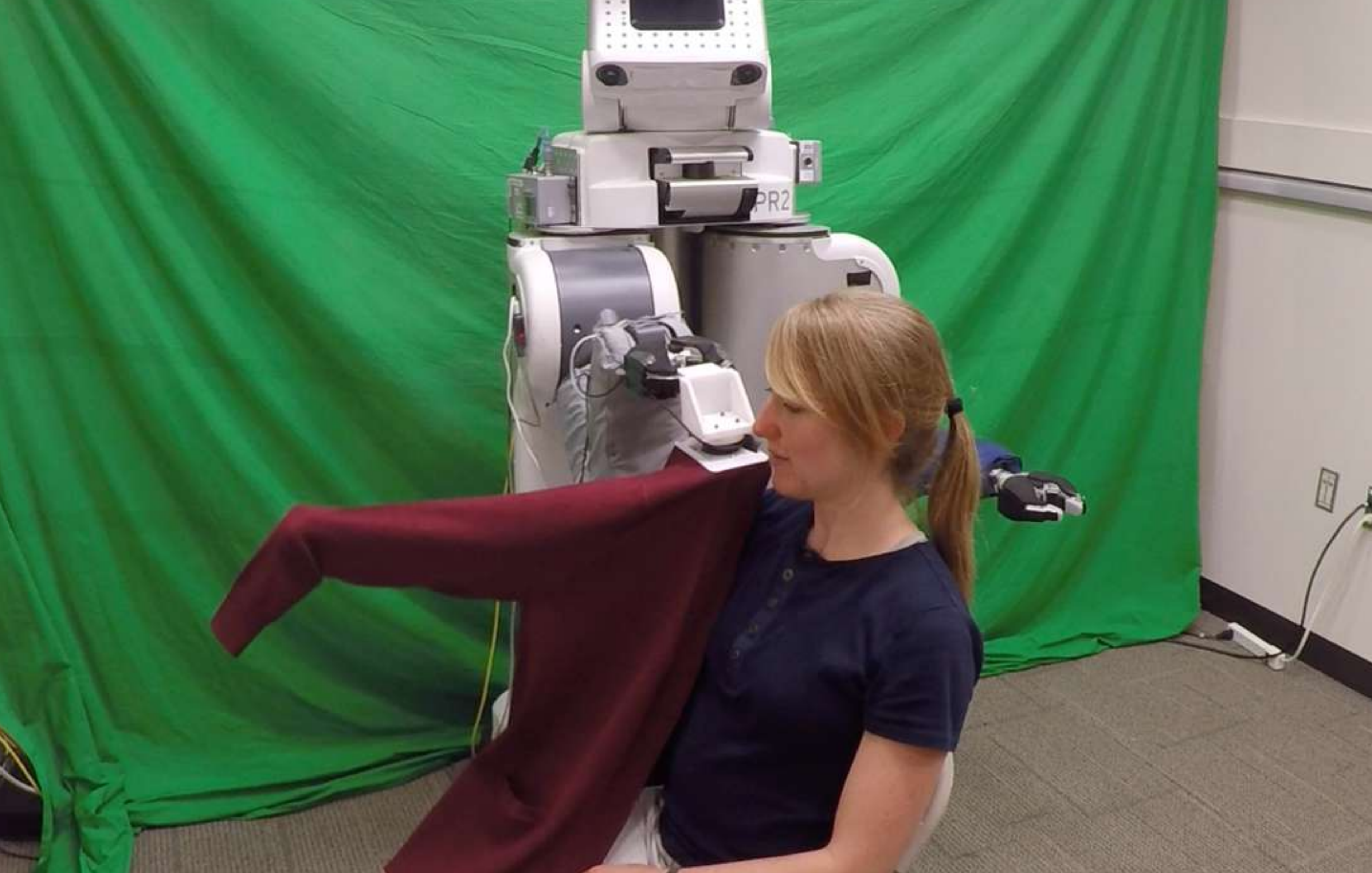}
\includegraphics[width=0.19\textwidth, trim={15cm 8cm 6cm 1cm}, clip]{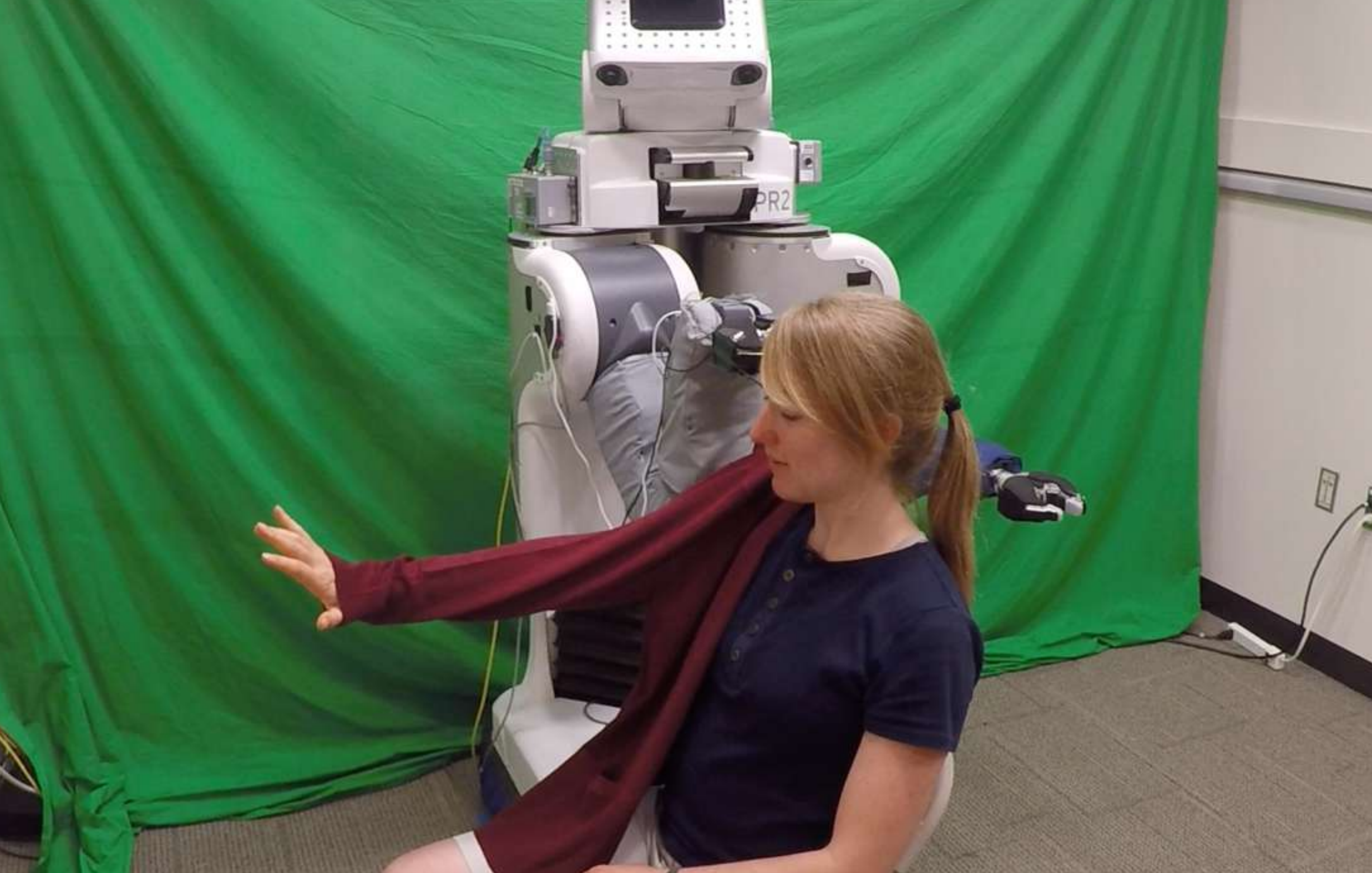}
\caption{\label{fig:cardigan}The PR2 pulls a long sleeve cardigan onto a participant's arm while tracking arm movement.}
\end{figure*}


\begin{figure}
\centering

\includegraphics[width=0.23\textwidth, trim={11.5cm 5cm 26.5cm 15cm}, clip]{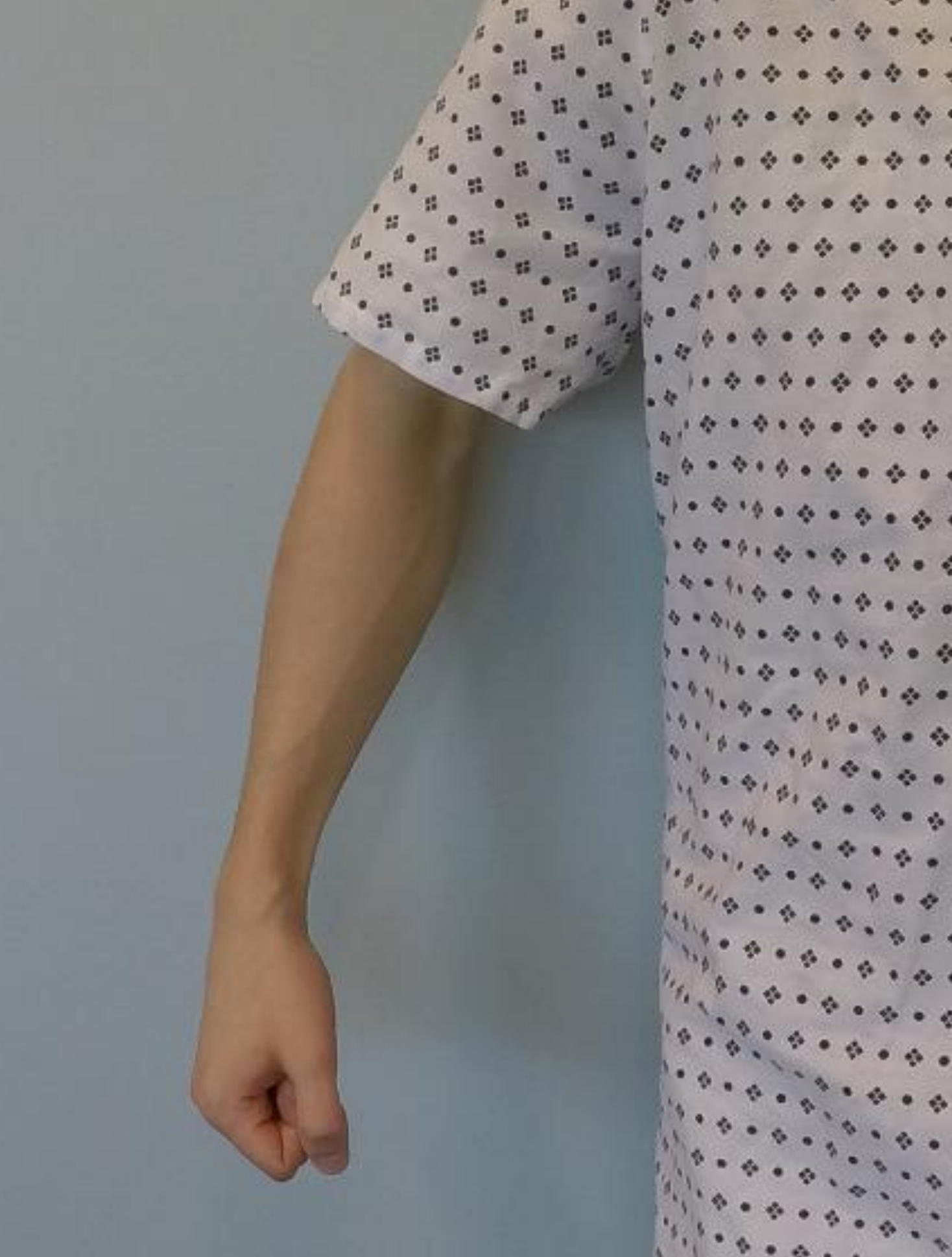}
\includegraphics[width=0.23\textwidth, trim={27cm 4.5cm 11cm 15.5cm}, clip]{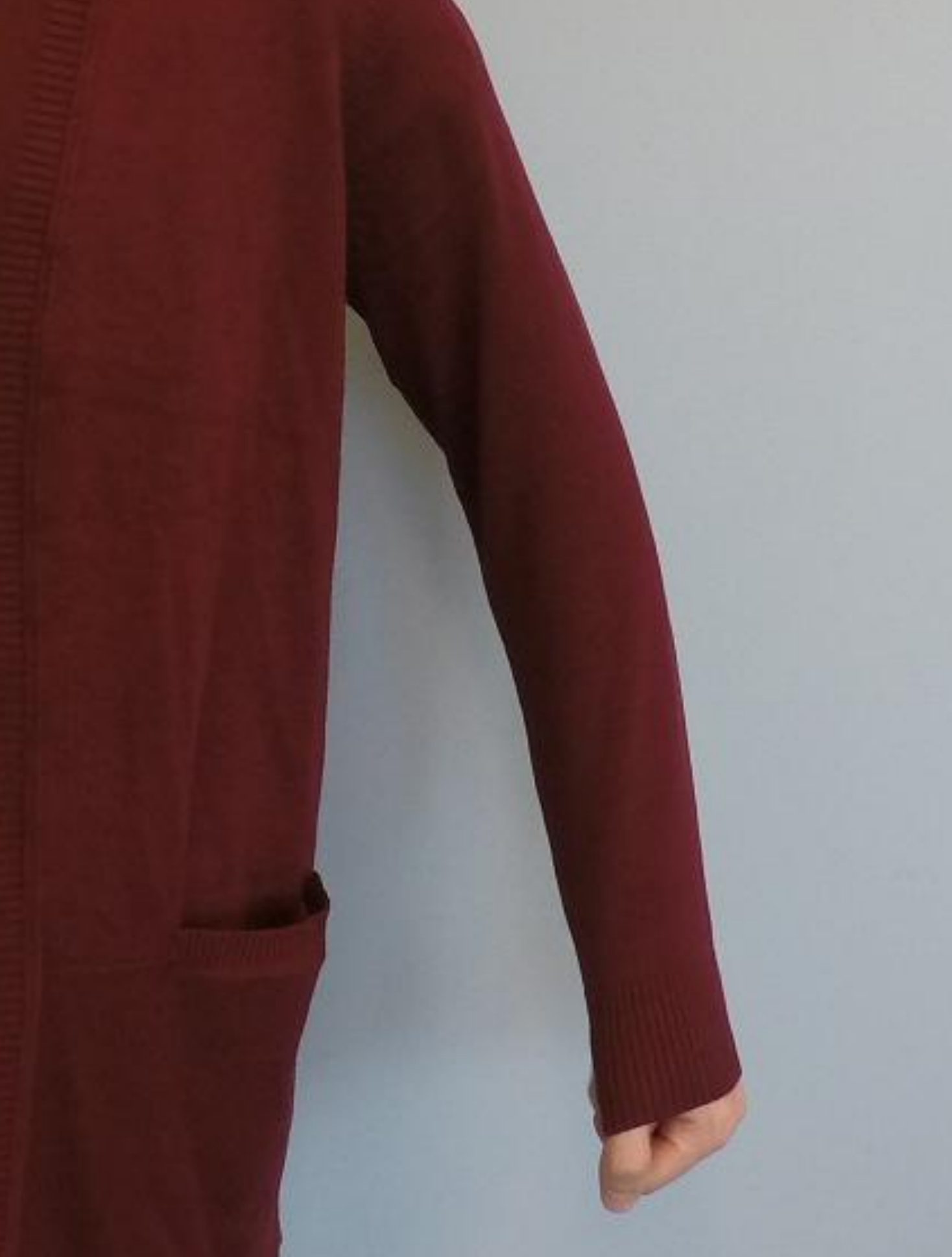}
\caption{\label{fig:gowncardigan}Sleeve differences between the hospital gown and knitted cardigan. The hospital gown has a short and wide sleeve, while the cardigan has a long and narrow sleeve.}
\end{figure}

\subsection{Adjusting to Arm Movement}
\label{sec:movement}

To illustrate our system's ability to adapt to human motion during robot-assisted dressing, we conducted three demonstrative trials in which each participant moved his or her arm vertically during dressing. We instructed participants to tilt their arms at any time during the dressing process such that their hands remained within 20~cm above or below the initial, horizontal pose. The robot's end effector started 5 cm above the person's hand and moved approximately parallel to the axis of their arm. Fig.~\ref{fig:demo} shows a series of images illustrating the movements of one of the participants and the PR2 adjusting to those movements. We observed that by using capacitive feedback control, the robot successfully pulled the gown onto every participant's arm, while tracking the vertical change in the person's arm pose during dressing. Fig.~\ref{fig:stage8} shows how the vertical position of the robot's end effector changes over time with respect to the estimated height of a participant's arm during a representative dressing trial. For clarity, Fig.~\ref{fig:stage8} displays only a single dressing trial, yet we note that these results remain consistent across trials from all 10 participants.

An additional feature of the capacitive sensor is its ability to detect contact between a person and the robot when $d(\Delta C)$ is arbitrarily close to zero. Given this, we can analyze any contact that occurred with the robot as each participant moved his or her arm during dressing. We detect contact when $d(\Delta C) < 0.5$~cm, which occurs when $\Delta C \geq 165$. During the trials with all 10 participants, the robot made contact with only two participants. For the first participant, contact occurred twice, the first lasting for $\sim$0.1s and the second lasting for $\sim$0.4s. The forces measured at the end effector during these collisions never exceeded 5 N. For the second participant, contact lasted $\sim$0.4s, yet no more than 0.5 N of force was measured. We note that even when collision occurs due to human motion, feedback control with capacitive sensing helps to ensure that contact duration is short and applied forces remain low.

During the next trial, we explored the capacitive sensor's ability to sense a person's body through clothing. For example, this may be applicable when helping dress a jacket onto someone who is already wearing a shirt. During this trial, each participant wore a long sleeve shirt that covered his or her arm. We again asked participants to vertically tilt their arms as the robot pulled on the sleeve of a hospital gown. Fig.~\ref{fig:stage9} shows how the vertical position of a participant's arm and the robot's end effector changes while the participant is wearing a long sleeve shirt. In addition, a series of images demonstrating this sequence can be seen in Fig.~\ref{fig:longsleeve}. Overall, we found that the results closely matched results from when participants' arms were not covered by a long sleeve garment. In both long and short sleeve scenarios, the robot's end effector remained near the target 5~cm range from a participant's arm with an average tracking error of $\sim$1.5~cm despite arm motion. The supplemental video further illustrates how our method can sense and track a person's arm movement, even when the person's arm is occluded by a long sleeve shirt. 

Finally, we evaluated how well our approach generalizes to dressing another garment, a long sleeve cardigan. We had the robot pull the long sleeve of the cardigan onto the right arm of all 10 participants, and we encouraged participants to vertically tilt their arms during dressing. The sequence in Fig.~\ref{fig:cardigan} displays the PR2 pulling the cardigan onto a participant's arm while tracking arm movement. Using our method, the PR2 was able to pull the cardigan fully up the arms of all 10 participants. In addition, the PR2's end effector tracked all arm movement, and contact between the end effector and a person's arm occurred only twice. For both occurrences, the contact duration was shorter than 0.4s and the end effector measured less than 2 N of applied force. 



\section{CONCLUSION}
In this work, we demonstrated that a simple capacitive sensor can be used to track human motion and adapt to pose estimation errors during robot-assisted dressing. With a capacitive sensor, our robot estimates the vertical distance between its end effector and a person's arm and we presented a closed form equation for estimating this distance, which enables these estimates to be made in real time with low computational resources. We showed that these capacitive sensors enable a robot to track vertical human motion using only a simple robot controller, and we detailed an approach to extend this to tracking 3D human motion.

A PR2 using our method was able to successfully dress the right arms of 10 human participants with a hospital gown, even when there were 15~cm of initial error in the estimated height of a person's arm. We show that this sensing approach is unaffected by visual occlusion and is able to sense the human body through fabric clothing, such as a long sleeve cotton shirt, which provides evidence that this approach may be useful in other scenarios for which humans and robots physically interact. We further demonstrated that our method allows a PR2 to track human motion and stay within 5~cm of a participant's arm while assisting with dressing both a hospital gown and a long sleeve cardigan (Fig.~\ref{fig:gowncardigan}). Finally, we found that capacitance measurements remained consistent with low variance across all participants, which provides evidence that this sensing approach may generalize well to a wide population of people.






\bibliographystyle{IEEEtran}

\end{document}